\documentclass[10pt,journal,compsoc]{IEEEtran}
\usepackage{bm}
\usepackage{algorithm}
\usepackage{algpseudocode}
\usepackage{enumitem}
\usepackage{amsmath}
\usepackage{amssymb}
\usepackage{multirow}
\usepackage{booktabs}
\usepackage{color}
\usepackage{makecell}
\usepackage{enumitem}
\usepackage{mathrsfs}

\usepackage{subfigure} 
\DeclareUnicodeCharacter{FB01}{fi}
\ifCLASSOPTIONcompsoc
  \usepackage[nocompress]{cite}
\else
  \usepackage{cite}
\fi

\ifCLASSINFOpdf
   \usepackage[pdftex]{graphicx}
\else
   \usepackage[dvips]{graphicx}

\fi

\usepackage{amsmath}

\usepackage{color}

\newcommand{\pihat}[0]{{\hat {\bm \pi}}}
\newcommand{\tauhat}[0]{{\hat {\bm \tau}}}
\newcommand{\mat}[1]{{\bf #1}}           %定义矩阵符号命令，\mat{X}
\renewcommand{\vec}[1]{{\boldsymbol #1}} %定义向量符号命令，\mat{x}
\usepackage{pifont}

\title{A Survey on Service Route and Time \\ Prediction in Instant Delivery: Taxonomy, \\ Progress, and Prospects} 

\author{Haomin~Wen,
	Youfang~Lin,	
        Lixia Wu,
	Xiaowei~Mao, 
        Tianyue~Cai, 
        Yunfeng~Hou, 
        Shengnan Guo, \\
        Yuxuan~Liang,
        Guangyin~Jin,
        Yiji Zhao,
        Roger~Zimmermann,
        Jieping Ye,
        Huaiyu~Wan*
    
        \IEEEcompsocitemizethanks{\IEEEcompsocthanksitem H. Wen, Y. Lin, X. Mao, T. Cai, Y. Hou, S. Guo, H. Wan are with the Beijing Key Laboratory of Traffic Data Analysis and Mining, School of Computer and Information Technology, Beijing Jiaotong University, Beijing 100044, China, and the Key Laboratory of Intelligent Passenger Service of Civil Aviation, CAAC, Beijing, 101318, China. E-mail: \{wenhaomin,yflin,maoxiaowei\}@bjtu.edu.cn; \\
        \{caitianyue, houyunfeng,guoshn,hywan\}@bjtu.edu.cn
	\IEEEcompsocthanksitem H. Wen, Z. Roger are with the School of Computing, National University of Singapore,  Singapore. E-mail: rogerz@comp.nus.edu.sg
       \IEEEcompsocthanksitem G. Jin is with College of Systems Engineering, National University of Defense Technology, Changsha, China. E-mail: jinguangyin96@foxmail.com
      \IEEEcompsocthanksitem Y. Zhao is with the School of Information Science and Engineering, Yunnan University, Kunming 650091, China. E-mail: yijizhaoneo@gmail.com
        \IEEEcompsocthanksitem L. Wu, X. Mao are with the Cainiao Network, Hangzhou, China. E-mail: wallace.wulx@cainiao.com
        \IEEEcompsocthanksitem J. Ye is with the Alibaba Group, Hangzhou, China. E-mail: yejieping.ye@alibaba-inc.com
        \IEEEcompsocthanksitem Y. Liang is with INTR Thrust and DSA Thrust, The Hong Kong University of Science and Technology (Guangzhou). E-mail: yuxliang@outlook.com
 }

\thanks{ \protect\\(Corresponding author: Huaiyu Wan.)}}

\makeatletter
\def\endthebibliography{%
	\def\@noitemerr{\@latex@warning{Empty `thebibliography' environment}}%
	\endlist
}
\makeatother
\begin{document}

\IEEEtitleabstractindextext{%
\begin{abstract}  
\par  Instant delivery services, such as food delivery and package delivery, have achieved explosive growth in recent years by providing customers with daily-life convenience. An emerging research area within these services is service Route\&Time Prediction (RTP), which aims to estimate the future service route as well as the arrival time of a given worker. As one of the most crucial tasks in those service platforms, RTP stands central to enhancing user satisfaction and trimming operational expenditures on these platforms. Despite a plethora of algorithms developed to date, there is no systematic, comprehensive survey to guide researchers in this domain. To fill this gap, our work presents the first comprehensive survey that methodically categorizes recent advances in service route and time prediction. We start by defining the RTP challenge and then delve into the metrics that are often employed. Following that, we scrutinize the existing RTP methodologies, presenting a novel taxonomy of them. We categorize these methods based on three criteria: (i) type of task, subdivided into only-route prediction, only-time prediction, and joint route\&time prediction; (ii) model architecture, which encompasses sequence-based and graph-based models; and (iii) learning paradigm, including Supervised Learning (SL) and Deep Reinforcement Learning (DRL). Conclusively, we highlight the limitations of current research and suggest prospective avenues. We believe that the taxonomy, progress, and prospects introduced in this paper can significantly promote the development of this field.

%Finally, we implement the mentioned methods and release a package for service route and time prediction to accelerate the research in this field. The source code is available at https://github.com/wenhaomin/route\_time\_prediction. Based on the implementation, we conducted extensive experiments on a real-world public dataset to benchmark the performance of different models.
\end{abstract}

\begin{IEEEkeywords}
{\ service route and time prediction, instant delivery;}
\end{IEEEkeywords}}

% make the title area
\maketitle
\IEEEdisplaynontitleabstractindextext
\IEEEpeerreviewmaketitle

\section{Introduction}
\par Instant delivery services \cite{dablanc2017rise, e_le_me, chen2022emerging}, such as logistics and food delivery, are playing an increasingly important role in serving people's daily demands. By the end of 2021, China's online food delivery platforms processed approximately 29.3 billion orders, engaging over 4 million workers and 460 million consumers. A crucial task on these service platforms is service route and time prediction (RTP) \cite{e_le_me, wen2021package, gao2021deep}, which aims to estimate the future service route and arrival time of a worker given his unfinished tasks. The RTP problem has received increasing attention from both academia and industry in recent years, as it is a foundation for building intelligent service platforms, such as the logistics platforms Cainiao\footnote{https://global.cainiao.com/}, JD.COM\footnote{https://www.jdl.com/}, and the food delivery platforms Meituan\footnote{https://www.meituan.com/}, GrabFood\footnote{https://food.grab.com/sg/en/}. For instance, accurate arrival time prediction can largely alleviate the waiting anxiety of customers \cite{gao2021deep,end2end2020Araujo, wu2019deepeta, DBLP:conf/kdd/WangFY18}, thus improving the customer's experience. Moreover, route forecasts can be integrated into dispatching systems, optimizing order assignments in proximity to a worker's route \cite{lan2020decomposition,wang2020intelligent,huang2020dynamic}. In light of the above benefits, precise RTP predictions not only elevate user experience but also reduce operational costs, therefore deserving further studies in the research community.

\par Thanks to the wide equipment of personal digital assistant (PDA) devices for workers, massive historical behaviors of workers are collected from their daily operations, such as GPS location, task accept-time, task finish-time, etc. This forms the data foundation for learning-based models to mine workers' behavior patterns,  particularly in terms of routing and estimated time of arrival, as we focus on in this paper. To this end, we have witnessed a variety of learning-based models dedicated to service route and time prediction in instant delivery recent years. However, there is no systematic and comprehensive survey to summarize and guide research in this field. This deficiency hinders researchers' grasp of both the present landscape and evolving trends of this research field.  Addressing this need, we introduce the first survey on RTP techniques,  offering a systematic overview and arrangement of the latest endeavors in this domain. Firstly, we define the RTP task and introduce commonly used metrics. Subsequently, we conduct an exhaustive examination of current RTP approaches, stratifying them across three criteria: (i) task perspective (only-route prediction, only-time prediction, both route and time prediction), (ii) model architecture (sequence-based, graph-based), and (iii) learning paradigm (supervised learning, deep reinforcement learning). The proposed taxonomy is shown in Figure~\ref{fig:Taxonomy}. At last, we discuss the limitations in current research and suggest potential directions for further exploration. Overall, we summarize our contributions as the following three points:

\begin{itemize}[leftmargin=*]
    \item \emph{The First Survey on RTP}: To the best of our knowledge, this is the first survey that encompasses a thorough examination of recent advancements in RTP research, ensuring a complete understanding of the field's progress and evolution.

    \item \emph{A Systematic Taxonomy and Classification}: We create a well-organized taxonomy and classification system for various RTP methods from three perspectives, enabling researchers to better comprehend the relationships between different approaches.

    \item \emph{Limitations and Future Directions}: We identify the limitations of current works and discuss the potential future research directions in route and time prediction, to inspire innovative ideas and promote growth within the domain.
\end{itemize}

%\par 3) \textbf{Comparative Analysis and Benchmarking}. Detailed comparative analysis and extensive benchmark experiments in a real-world logistics dataset are conducted, focusing on the strengths and weaknesses of different RTP algorithms to provide valuable insights into their applicability and effectiveness.

%\par 4) \textbf{Release of RTP package.} We implement different route and time prediction algorithms and release a Python package, to facilitate research in this field. The code can be found at https://github.com/wenhaomin/route\_time\_prediction.

\par \textbf{Comparisons to Existing Surveys}. Since there are no directly related surveys, we compare our work with existing literature in two key areas: route-related surveys and time-related surveys. Firstly, one direction of route-related surveys primarily concentrates on route optimization \cite{amin2021survey, caceres2014rich, ichoua12007planned, goel2017vehicle}. These studies aim to plan the best route for workers based on metrics like travel distance. In contrast, our work centers on route prediction, seeking to forecast the route a worker is most likely to choose. Additionally, another avenue in route-related surveys addresses the next location prediction problem \cite{sabarish2015survey, wu2018location, zheng2018survey, xu2020survey, koolwal2020comprehensive}, aiming to select the most probable next location a user will visit from a set of common location candidates (i.e., all predictions share the same location candidates). Unlike those studies, the predicted route in RTP is constrained by the unfished tasks.  Specifically, the location candidates in the estimated route are derived from, and vary with, the input of incomplete tasks (i.e., different predictions have different candidates), making it a more challenging problem to tackle.
Thirdly, earlier time-related surveys focus on predicting arrival times \cite{altinkaya2013urban,oh2018short,reich2019survey,singh2022review} in the map system, particularly for buses. Our work, however, targets arrival time prediction in instant delivery, a more complex task due to the challenge of predicting workers' actions. In summary, our research addresses the RTP problem in the emerging field of instant delivery, an area that deserves further investigation.

%Our survey diverges from existing literature in several key aspects: 
%i) Novel topic, unlike previous surveys focusing on problems that have been studied for years (such as next location prediction \cite{wu2018location, koolwal2020comprehensive, zheng2018survey} or time prediction in the map system), we focus on the emerging RTP problem raised in instant delivery.

%Although there are surveys on trajectory data analysis or ETA prediction in the map system, no relevant surveys on RTP prediction in instant delivery exist.  

\par \textbf{Organization.} The survey is structured as follows: Section \ref{sec:preliminaries} formulates the problem and introduces the frequently employed metrics. Section~\ref{sec:taxonomy} presents the proposed taxonomy. Section~\ref{sec:route_prediction}-\ref{sec:route_and_time_prediction} introduces the details of current route prediction, time prediction, route and time prediction methods, respectively. Lastly, Section~\ref{sec:limitation_future_direction}-\ref{sec:conclusion} makes the conclusion which analyzes the limitations and points out the future directions in this research field.
% Section~\ref{sec:benchmark_experiments} provides extensive benchmark results for different methods.

\section{Preliminaries} \label{sec:preliminaries}

\par We first give a general formulation of RTP to facilitate the following sections (Different methods can have different formulations which are slightly different from the general form, as we will introduce later). Then we introduce commonly used metrics for evaluating different methods.

\par
\begin{figure}[hbtp]
    \centering
    \includegraphics[width=0.7 \columnwidth]{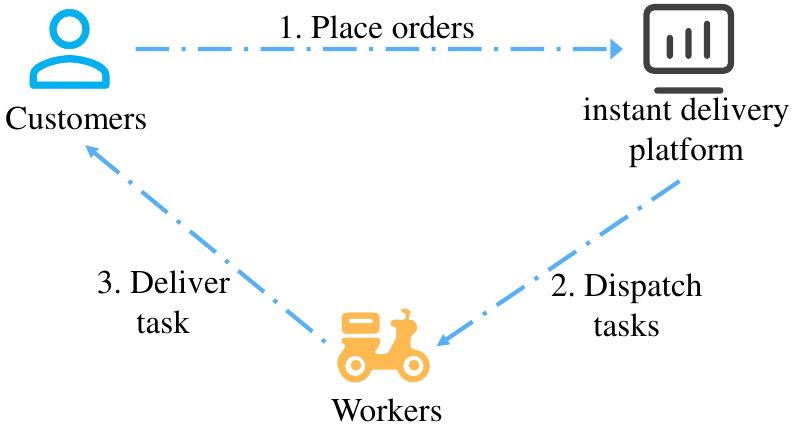}
    
    \caption{Illustration of instant delivery service.}
    \label{fig:background}
\end{figure}

 \noindent \textbf{Background.} As shown in Figure~\ref{fig:background}, in a typical instant delivery service, such as food delivery, the customer first places a task (order) with certain spatial-temporal requirements such as delivery location and required time window (i.e., 9:00 am - 9:30 am) in the online platform, then the platform will dispatch the task to a worker. At last, the worker will try to finish the task while satisfying the time and location requirements. 

\par
\begin{figure}[hbtp]
    \centering
    \vspace{-0.5em}
    \includegraphics[width=1 \columnwidth]{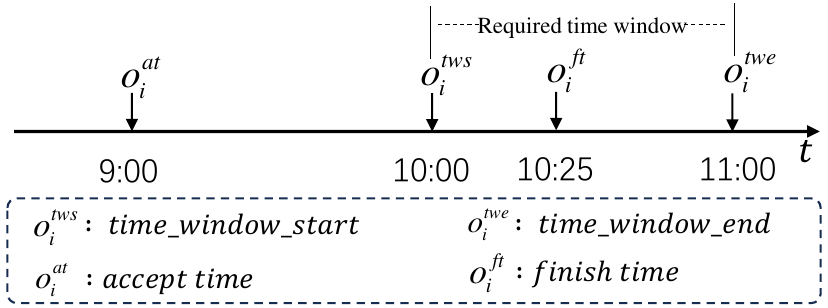}
    \vspace{-1em}
    \caption{Illustration for the time features of task $o_i$.}
  
    \label{fig:task_time}
\end{figure}

 \noindent \textbf{Definition 1: Task.} A task represents a pick-up or a delivery order in the platform. Different services have different types of tasks. For instance, there are only pick-up tasks in the package pick-up service, while both pick-up and delivery tasks exist in the food delivery service. Given a task denoted by $o_i$, it is associated with both spatial and temporal features:
\begin{equation}
    {\vec o}_i = (o_i^{lat}, o_i^{lng}, o_i^{aoi}, o_i^{type}, o_i^{at},  o_i^{ft}, o_i^{tws}, o_i^{twe}),
\end{equation}
where the spatial features of task $o_i$ include:
\begin{itemize}[leftmargin=*]
    \item $o_i^{lat}$,  the latitude of the task; 
    \vspace{0.2em}
    \item $o_i^{lng}$,  the longitude of the task;
    \vspace{0.2em}
    \item $o_i^{aoi}$ (if applicable),  the ID of the Area-Of-Interest (AOI) where the task is located in.
    \vspace{0.2em}
    \item $o_i^{type}$ (if applicable), the type (e.g., school area, residential area) of the task's AOI. 
    %It can be obtained from the GaoDe\footnote{https://lbs.amap.com/} platform by taking the task location as a query.
\end{itemize}
And the temporal features of task $o_i$ include:
\begin{itemize}[leftmargin=*]
    \item $o_i^{at}$,  the accept-time (by worker) of the task.
    \vspace{0.2em}
    \item $o_i^{ft}$,  the finish-time of the task.
    \vspace{0.2em}
    \item $o_i^{tws}$,  the start of the required time window.
    \vspace{0.2em}
    \item $o_i^{twe}$,  the end of the required time window.
\end{itemize}
In Figure~\ref{fig:task_time}, we further give a simple illustration of the time features related to task $o_i$ to facilitate the understanding.

\par \noindent \textbf{Definition 2: Worker.} A worker $w$ is responsible for the tasks generated by customers. For instance, in the logistics platform, the worker is the courier for picking up/delivering packages. While in the food delivery platform, the worker is the courier that needs to finish both pickup and delivery tasks. Each worker is associated with his personalized features $\vec x_w$, such as the total number of work days in the platform, the average number of tasks in a day, the average arrival time, etc.

\subsection{General Service Route\&Time Prediction Problem}

% sequence
\par \noindent \textbf{Definition 3: Finished Tasks.} At a certain time $t$, a worker $w$ has $m$ finished tasks, denoted by $\mathcal{O}_{w,t}^{f} = <o_1, \dots, o_m>$, where $o_i^{ft} \le t$ for 
$i \in \{1,\dots,m\}$. It is worth mentioning that $\mathcal{O}_{w,t}^{f}$ is essentially a sequence sorted by the task's finish time, i.e.,  $o_i^{ft} \le o_j^{ft}$ if $i < j$. For simplicity, we remove the subscript $w$ of the variables related to a worker, such as $\mathcal{O}_t^{f}$ in the following description.

\par \noindent \textbf{Definition 4: Unfinished Tasks.} At time $t$, a worker $w$ can also have $n$ unfinished tasks, denoted by $\mathcal{O}_t^{u} = \{o_1, \dots, o_n\}$, where $o_i^{at} \le t \le o_i^{ft}$ (i.e., accepted but not finished) for $i \in \{1,\dots,n\}$.  Unlike the finished tasks, we consider $\mathcal{O}_t^{u}$ as a set since the finish time of each task $\mathcal{O}_t^{u}$ in is not known.

\par Since the unfinished tasks is a prerequisite for all route and time prediction models, here we introduce some commonly used features by current models that can influence the couriers' routing decisions and arrival time. Each task $o_i$ is associated with a feature ${\vec x_i}$, which can be divided into time-invariant ${\vec x}_i^{ti}$ and time-variant ${\vec x}_i^{tv}$ features. Specifically, the time-invariant features of the $i$-th unfinished task include: 
\begin{equation}
    {\vec x}_i^{ti} = (o_i^{lat}, o_i^{lng}, o_i^{aoi}, o_i^{type}, o_i^{at},  o_i^{tws}, o_i^{twe}).
\end{equation}
Where each feature has been explained in the definition of a task, and the time-variant feature includes:
\begin{equation}
    {\vec x}_i^{tv} = (o_i^{d}, o_i^{tws}-t, o_i^{twe}-t, t-o_{i}^{at}), 
\end{equation}
\begin{itemize}[leftmargin=*]
    \item $o_i^{d}$ is the distance between the task and the worker's current location, since workers tend to visit the nearby task first.
    \item $o_i^{tws}-t / o_i^{twe}-t$ calculates the time duration between the current time and the required time window. Workers tend to visit the more urgent task first.
    \item $t-o_{i}^{at}$, which is the duration that the task has joined the worker's task pool. The longer a task is in a worker's task pool, the more likely it will be visited next by the worker.
\end{itemize}

% Different types of locations have different patterns, such as the average task number. Workers can have preferences on different types of locations.

\par \noindent \textbf{Definition 5: Route Constraints.} In reality, various route constraints can exist in different services, such as the pick-up then delivery constraint (i.e., the delivery location of an order can only be visited after its pick-up location is visited \cite{gao2021deep, parragh2008survey}) and capacity constraint (i.e., the total weight of items carried by a worker can not exceed its capacity of load \cite{toth2002vehicle, augerat1998separating}). Route constraints of a problem can be represented by a rule set $\mathcal C$, with each item corresponding to a specific route constraint. 

\par \noindent \textbf{Definition 6: Route Prediction Problem.} Given a worker $w$'s finished and unfinished tasks at time $t$, route prediction aims to learn a mapping function ${\mathcal F}_{R}$ to predict the worker's future service route $\hat {\bm \pi}$ of unfinished tasks which can satisfy the given route constraints $\mathcal C$, formulated as:

\begin{equation}
    {{\mathcal F}_{R}}(\mathcal{O}_t^{f}, \mathcal{O}_t^{u};{\mathcal C}) \rightarrow \pihat = [{\hat \pi _1},{\hat \pi _2} \cdots {\hat \pi _{n}}], \\
\end{equation}
where $\pihat$ is essentially a permutation of the unfished tasks $\mathcal{O}_t^{u}$, where $\hat \pi_i$ means that the $i$-th node in the route is task ${\hat \pi_i}$. Moreover, ${\hat \pi _i} \in \{ 1, \cdots n\} \; {\rm and}\;{\hat \pi _i} \ne {\hat \pi _j}\;{\rm if}\;i \ne j$.

\par \noindent \textbf{Definition 7: Time Prediction Problem.} Given a worker $w$'s finished and unfinished tasks at time $t$, time prediction aims to learn a mapping function ${\mathcal F}_{T}$ to predict the worker's arrival time for all unfinished tasks, formulated as: 
\begin{equation}
    {{\mathcal F}_{T}}(\mathcal{O}_t^{f}, \mathcal{O}_t^{u}) \rightarrow \tauhat = [{\hat \tau _1},{\hat \tau _2} \cdots {\hat \tau _{n}}], \\
\end{equation}
where ${\hat \tau_i} = t_{i}^{ft} - t$ means how long the worker can arrive at (or finish) the task $i$ since the query time $t$. 

\par \noindent \textbf{Definition 8: Route\&Time Prediction Problem.} Similarly, we formulate the route and time prediction problem as follows:
\begin{equation}
    {{\mathcal F}_{\mathcal RT}}(\mathcal{O}_t^{f}, \mathcal{O}_t^{u};{\mathcal C}) \rightarrow (\pihat, \tauhat). \\
\end{equation}

\begin{figure}[hbtp]
    \centering
    \includegraphics[width=1 \columnwidth]{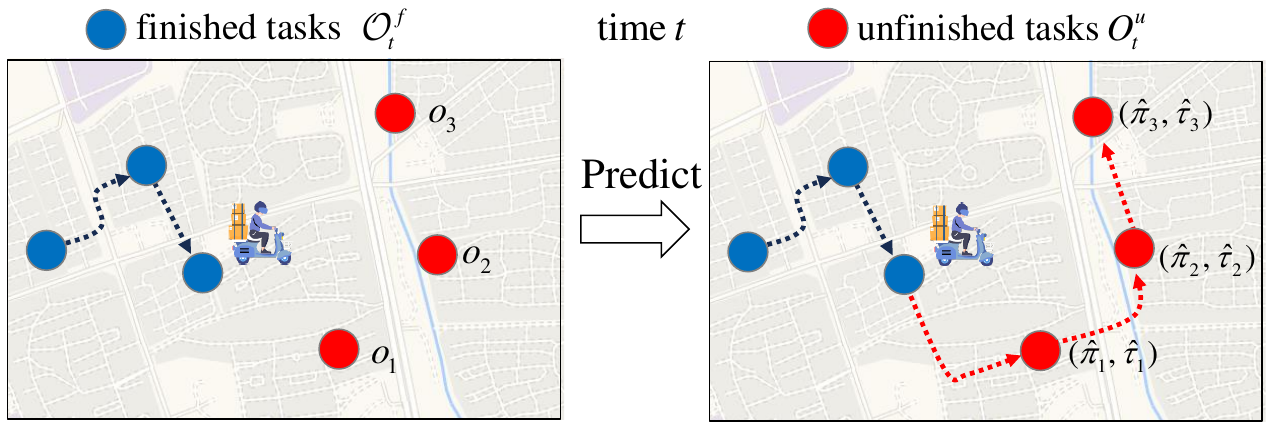}
    \caption{Illustration of the route and time prediction problem. }
    \label{fig:rtp_problem}
\end{figure}

We give an illustration of the route and time prediction problem in Figure~\ref{fig:rtp_problem}. And Table~\ref{tab:notation} lists all the related notions in the paper. 

\begin{table}[htbp]
	\caption{Summary of symbol notations.}
	\small
	\begin{center}
		\renewcommand\arraystretch{1.2}
		\begin{tabular}{|c|p{0.75 \linewidth}|}
			
			\hline
			\multicolumn{1}{|c|}{\textbf{Notation}} & \multicolumn{1}{|c|}{\textbf{Definition}} \\ 
			\hline
                $w$  & the target worker  \\  \hline
                $\mathcal C$ & route constraints \\ \hline
                %$\mathbf{X}_t^{w}$ & Features of $w$'s unfinished tasks at time $t$\\ \hline
                ${\mathcal O}_{t}^{f}$ & worker $w$'s finished tasks at time $t$ \\ \hline
                ${\mathcal O}_{t}^{u}$ &  worker $w$'s unfinished tasks at time $t$ \\ \hline
                $\pihat$  & $\pihat = \{{\hat \pi}_1, \dots, {\hat \pi}_n\}$, predicted service route\\  \hline
               
                ${\bm \pi}$  & ${\bm \pi} = \{\pi_1, \dots, \pi_{n^{\prime}}\}$,  the route label \\  \hline
                $n^{\prime}$ & number of tasks in the route label\\ \hline
                
                $Y_{\bm \pi}(i)$ & the order of task $i$ in the label route \\ \hline
                $Y_{\bm \pihat}(i)$ & the order of task $i$ in the predicted route \\ \hline
                ${\bm \tau}/ {\bm {\hat \tau}}$  &  actual / predicted arrival time\\  \hline
                
                $\mat E$ & the embedd matrix of all unfinished tasks\\ \hline
                $\vec h_j$ & the hidden state of decoding step $j$\\ \hline
                $u_i^j$ & the compatibility score of $i$ at decoding step $j$ \\ \hline

                \multicolumn{2}{|c|}{RL-related notation} \\ \hline
                 $M$ & the Markov Decision Process\\ \hline
                 $S$ & the set of states \\ \hline
                 $A$ & the set of actions\\ \hline
                 $P$ & the transition probability\\ \hline
                 $R$ & the reward function\\ \hline
                 $\gamma$ & the discount factor\\ \hline

                \multicolumn{2}{|c|}{Graph-related notation} \\ \hline
                
			$\mathcal{G}_t^{w}$  & Input ST-Graph of worker $w$ at time $t$  \\  \hline
			%$\mathcal{V}_t^F$ / $\mathcal{V}_t^U$  &  Finished / Unfinished nodes at time $t$   \\  \hline
			
			$\mathcal{N}_i$ & Neighbors of node $i$ \\  \hline
			
			${\mat X}_t^{v}$ / ${\mat X}_t^{e}$ & Node / Edge features at time $t$  \\\hline
			${\mat E}_t$ / ${\mathbf Z}_{t}$  & Node / Edge embeddings after encoding  \\\hline
		
		\end{tabular}
		\label{tab:notation}
	\end{center}
\end{table}

\subsection{Metrics}
\par Here, we introduce a comprehensive metric system to evaluate the performance of route prediction and time prediction, respectively. 

\subsubsection{Evaluation of Route Predictioin}

\par Note that in some instant delivery services (e.g., food delivery), the tasks of a worker are not settled from the beginning. Rather, they are revealed over time because the platform can continuously dispatch new tasks to the worker. In that case, the new task coming at $t^{\prime}$ can change the worker's previous decisions at $t$, making observations after $t^{\prime}$ inaccurate \cite{wen2021package, gao2021deep} as the label for the sample at time $t$. Therefore, a better way is to treat the route observations between $t$ and $t^{\prime}$ as the label information when training or evaluation, recall that $t^{\prime}$ is the dispatch time of the first coming task after $t$. 
% Take Figure~\ref{fig:model} as an example, where the task's finish- or dispatch-event of a worker are depicted in the timeline. For the sample at $t_1$, its input contains task $\{1,2,3\}$, and the label ${\bm \pi}_{t_1:t_1^{'}}={\bm \pi}_{t_1:t_2}=[2]$.

At the evaluation process, formally, we have the prediction $\hat{\bm \pi} = [{\hat{\pi}_1}, \dots, {\hat{\pi}_n} ]$ and the label ${\bm \pi} = [\pi_1, \dots, \pi_{n^{\prime}} ]$, where $n^{\prime} \leq n ~ {\rm and} ~ {\rm set}({\bm \pi}) \subseteq {\rm set}({\hat {\bm \pi}})$. Let $Y_{\bm \pi}(i)$ and $Y_{\hat {\bm \pi}}(i)$ be the order of node $i$ in the label and prediction route, respectively. One can evaluate the route prediction performance by the following metrics from both global and local perspectives.

\vspace{0.5em}
\par \noindent \textbf{From the Global Perspective}. Metrics in this line measure the overall similarity of two input sequences, including:
\vspace{0.5em}

\begin{itemize}[leftmargin=*]
    \item \textbf{KRC}: Kendall Rank Correlation \cite{kendall1938new} is a statistical criterion to measure the ordinal association between two sequences. Given any task pair $(i, j)$, it is said to be concordant if both $Y_{\hat {\bm  \pi}}(i) > Y_{\hat {\bm  \pi}}(j)$ and $Y_{\bm \pi}(i) > Y_{\bm \pi}(j)$ or both $Y_{\hat {\bm  \pi}}(i) < Y_{\hat {\bm  \pi}}(j)$ and $Y_{\bm \pi}(i) < Y_{\bm \pi}(j)$. Otherwise, it is said to be discordant. To calculate this metric, tasks in the prediction are first divided into two sets: i) tasks in label ${{\mathcal O}_{in}} = \{ {\hat \pi}_i | {\hat \pi}_i \in {\bm \pi} \}$, and ii) tasks not in label ${{\mathcal O}_{not}} = \{ {\hat \pi}_i |  {\hat \pi}_i \not \in {\bm \pi} \}$. We know the order of items in $\mathcal{O}_{in}$, but it is hard to tell the order of items in  ${\mathcal O}_{not}$, still we know that the order of all items in $\mathcal{O}_{in}$ are ahead of that in ${\mathcal O}_{not}$. Therefore, KRC compares the task pairs $\{(i,j) | i,j \in {\mathcal O}_{in}~{\rm and}~{i \neq j}\} \cup \{(i,j) | i \in {\mathcal O}_{in} {~\rm and~~} j \in {\mathcal O}_{not} \}$. In this way, it is defined as:
    \begin{equation}
	{\rm{KRC}} = \frac{N_c-N_d}{N_c+N_d},
	\label{eq:krc}
    \end{equation}
    where $N_c$ is the number of concordant pairs, and $N_d$ is the number of discordant pairs.

    \item \textbf{ED:} Edit Distance \cite{nerbonne1999edit} (ED) is an indicator to quantify the dissimilarity of two sequences, by counting the minimum number of required operations to transform one sequence (in this case, the route prediction) into another (i.e., the actual route), formulated as:
    \begin{equation}
        {\rm ED} = {\rm EditDistance}({\overline {\bm \pi}}, {\bm \pi}).
    \end{equation}
    %where $\overline {\bm \pi} = [i ~{\rm for}~ i ~{\rm in}~ {\bm \pihat}\ ~{\rm if}~ i ~{\rm in}~ {\bm \pi}]$, 
    where $\overline {\bm \pi} = {\pihat} \cap {\bm \pi}$, 
    which is the common part of the prediction and label, with items' relative order in the prediction preserved.

    \item \textbf{LSD} and \textbf{LMD} \cite{wen2021package}: The Location Square Deviation (LSD) and the Location Mean Deviation (LMD) measure the degree that the prediction deviates from the label, formulated as:
    \begin{equation}
        \begin{aligned}
            {\rm LSD}&=\frac{1}{n^{\prime}}\sum_{i=1}^{n^{\prime}}(Y_{\pi}(\pi_i)-Y_{\hat \pi}(\pi_i))^2 \\
        {\rm LMD}&=\frac{1}{n^{\prime}}\sum_{i=1}^{n^{\prime}}|Y_{\pi}(\pi_i)-Y_{\hat \pi}(\pi_i)|. \\
        \end{aligned}
        \label{eq:lsd_lmd}
    \end{equation}

\end{itemize}

\begin{itemize}[leftmargin=*]
    \item \textbf{DMAE} \cite{ILRoute2023feng}: It denotes the mean absolute error of the distance differences between generated routes and real routes. It measures how far the generated routes deviate from the real routes in terms of spatial distance.
     \begin{equation}
            {\rm DMAE}=\frac{1}{n^{\prime}}\sum_{i=1}^{n^{\prime}}|{\rm Distance}(\hat{\pi}_i, \pi_i)|,
    \end{equation}
    where ${\rm Distance}(\cdot)$ is the distance function, which calculates the distance given two tasks.
  
    \item \textbf{SR@$k$} \cite{ILRoute2023feng}: which represents the relaxed concordancy rate of generated routes compared with real routes. It first calculates the distance between nodes in the generated route and the real route. If the distance is less than $k$ meters, the two route nodes are considered to be consistent. The number of consistent nodes is then counted, and this count is divided by the length of the routes to obtain the metric. The purpose of relaxing the distance criteria is to account for statistical errors that may arise when workers visit tasks from the same location, formulated as:
    \begin{equation}
         {\rm SR}@k= \frac{1}{n^{\prime}}\sum_{i=1}^{n^{\prime}}{\mathbb{I}}(|{\rm Distance}(\hat{\pi}_i, \pi_i)|<k),
    \end{equation}
    where ${\mathbb I}(\cdot)$ is the indicator function, and ${\mathbb{I}}(|{\rm Distance}(\hat{\pi}_i, \pi_i)|<k)$ equals 1 if  $|{\rm Distance}(\hat{\pi}_i, \pi_i)|<k$ else 0.
    
    \item \textbf{MRR} \cite{gao2021deep}: The Mean Reciprocal Rank measures whether the model can predict the actual next location with a higher probability, calculated by averaging the reciprocal of the actual locations’ ranks:
    \begin{equation}
        {\rm MRR}=\frac{1}{n^{\prime}}\sum_{i=1}^{n^{\prime}}\frac{1}{|(Y_{\pi}(\pi_i)-Y_{\hat \pi}(\pi_i))|+1}.
    \end{equation}
\end{itemize}

\vspace{0.5em}
\par \noindent \textbf{From the Local Perspective}. Metrics in this line focus on evaluating the performance of top-$k$ prediction, including:
\vspace{0.5em}

\begin{itemize}[leftmargin=*]
    \item \textbf{HR@$k$} \cite{wen2021package}: Hit-Rate@$k$ is used to quantify the similarity between the top-$k$ items of two sequences. It describes how many of the first $k$ predictions are in the label, which is formulated as follows:
    \begin{equation}
        \textbf{\rm HR@}k=\frac{|{\hat {\bm \pi}}_{[1:k]}\cap {\bm \pi}_{[1:k]}|}{k},
        \label{eq_hit_rate}
    \end{equation}
    where $|\cdot|$ means the cardinality of a set.

    \item \textbf{Same@$k$} \cite{graph2route}: Compared with HR@$k$, Same@$k$ is a more strict measurement to calculate the local similarity of two sequences. It answers the following question: Is the route composed of the first $k$ predictions exactly the same as the label? 
    \begin{equation}
        \textbf{\rm Same@}k= {\prod_{i=0}^k{\mathbb{I}}({\hat \pi}_i, \pi_i)}, 
    \end{equation}
    where ${\mathbb I}(\cdot)$ is a indicator function, and $\mathbb I({\hat \pi}_i, \pi_i)$ equals 1 if ${\hat \pi}_i$ equals $\pi_i$ else 0.
\end{itemize}

\par In summary, KRC, ED, LSD, LMD, and MRR measure the overall similarity of the predicted route and the label route according to tasks' orders in the two sequences. And DMAE, SR@$k$  measures the overall similarity based on the task's distance in the geographical location. In comparison, HR@$k$ and Same@$k$ calculate their similarity from the local perspective. Higher KRC, HR@$k$, Same@$k$, SR@$k$, MRR, and lower ED, LSD, LMD, DMAE mean better performance of the algorithm.

\subsubsection{Evaluation of Time Predictioin}

\par Time prediction is typically regarded as a regression problem. Thus metrics for the regression problem are employed to evaluate the performance. Let $\tau_i$, ${\hat \tau}_i$ be the actual arrival time and the predicted arrival time, respectively. And $n$ is the total number of unfinished tasks. The following metrics can be used:

\begin{itemize}[leftmargin=*]
    \item \textbf{MAE} \cite{wu2019deepeta}. Mean Absolute Error (MAE) is a commonly used metric, formulated as follows:
    \begin{equation}
      {\rm MAE}=\frac{1}{n} \sum_{i=1}^{n}\left|\hat{\tau}_{i}-{\tau}_{i}\right|.
    \end{equation}
    % \begin{equation}
    %     \renewcommand{\arraystretch}{2}
    %     \begin{array}{l}
    %     {\rm RMSE}=\sqrt{\frac{1}{N} \sum_{i=1}^{N}\left(\hat{\tau}_{i}-{\tau}_{i}\right)^{2}} \\
    %     {\rm MAE}=\frac{1}{N} \sum_{i=1}^{N}\left|\hat{\tau}_{i}-{\tau}_{i}\right|, \\
    %     \end{array}  
    % \end{equation}

    \item \textbf{RMSE} \cite{wu2019deepeta}. Root Mean Squared Error (RMSE) is another commonly used metric:
    \begin{equation}
      {\rm RMSE}=\sqrt{\frac{1}{n} \sum_{i=1}^{n}\left(\hat{\tau}_{i}-{\tau}_{i}\right)^{2}}. \\
    \end{equation}

    \item \textbf{MAPE} \cite{gao2021deep}. Mean Absolute Percentage Error, formulated as:
    \begin{equation}
      {\rm MAPE}=\frac{1}{n} \sum_{i=1}^{n}\left|\frac{\hat{\tau}_{i}-{\tau}_{i}}{\tau_i}\right|.
    \end{equation}

    \item \textbf{ACC@$k$} \cite{qiang2023i2rtp}. Besides the above traditional metrics. Delivery platforms usually provide an interval of arrival time for customer notification. Thus ${\rm ACC}@k$ is introduced by computing the ratio of prediction where the time difference between predicted time and true time is less than $k$ minutes, formulated as
    \begin{equation}
         {\rm ACC}@k= \frac{1}{n}\sum_{i=1}^n{\mathbb{I}}(|\hat{\tau_i}-\tau_i|<k).
    \end{equation}
    where ${\mathbb I}(\cdot)$ is the indicator function, and $\mathbb I({\hat \pi}_i, \pi_i)$ equals 1 if  $|\hat{\tau_i}-\tau_i|<k$ else 0. Usually, ACC@$30$ is adopted to test the model's ability in one-hour prediction.

\end{itemize}

\par Overall, route prediction metrics focus on evaluating the similarity between two ranked sequences, while time prediction metrics evaluate the regression error between predictions and labels.

\section{The proposed taxonomy} \label{sec:taxonomy}
\par
\begin{figure}[!t]
    \centering
    \includegraphics[width=1 \linewidth]{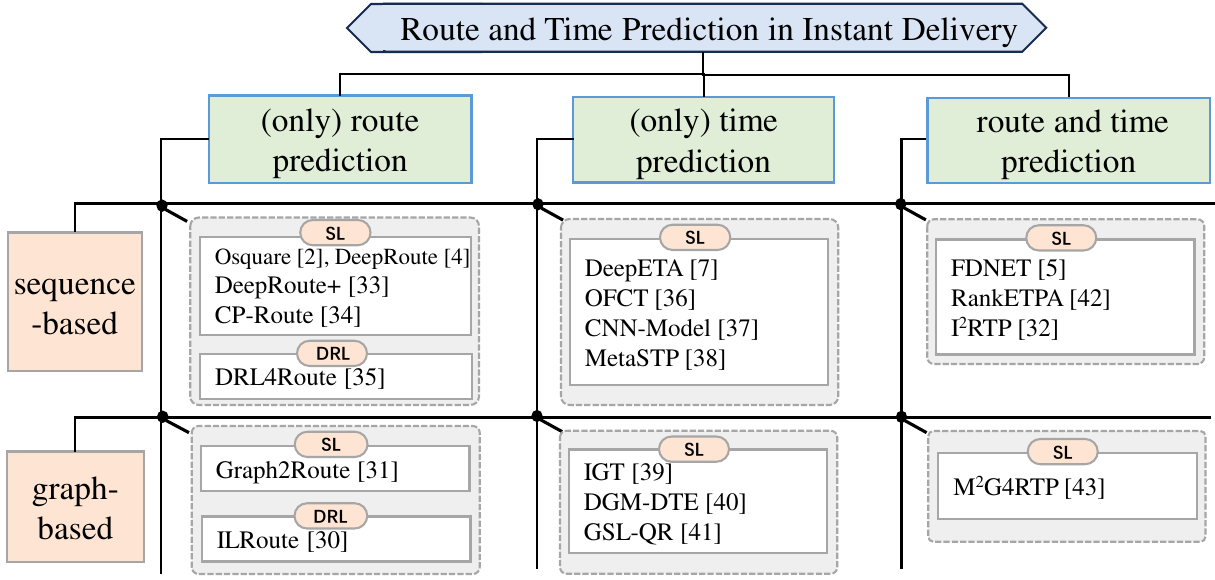}
    \caption{The proposed taxonomy of RTP algorithms for instant delivery. We summarize these methods from three dimensions, (i) from the task perspective, which has three categories: only-route prediction, only-time prediction, and route\&time prediction. (ii) from the perspective of model architecture, including sequence-based and graph-based models; (iii) from the perspective of learning paradigm: Supervised Learning (SL) and Deep Reinforcement Learning (DRL).}
    \label{fig:Taxonomy}
\end{figure}

 This paper provides a comprehensive review of current state-of-the-art models for the RTP task. In this section, we introduce the overall taxonomy of existing efforts, which is shown in Figure~\ref{fig:Taxonomy}. And the summary for neural architectures of existing route and time prediction models is shown in Table~\ref{tab:all_models}.

\par In the proposed taxonomy, we classify the existing methods by three dimensions: (i) task type (including route prediction, time prediction, and route\&time prediction); (ii) model architecture (sequence-based and graph-based); (iii) learning paradigm (supervised learning, deep reinforcement learning). Here we briefly introduce each classification dimension.

\subsection{Task Type}
\par  Broadly speaking, existing algorithms fall into three categories according to their task type, including:
\begin{itemize}[leftmargin=*]
    \item \textbf{(Only) route prediction}. Models in this category only aim to solve the route prediction problem, including Osquare \cite{e_le_me}, DeepRoute \cite{wen2021package}, DeepRoute+ \cite{DeepRoute+}, CP-Route \cite{wen2023modeling}, Graph2Route \cite{graph2route}, DRL4Route \cite{drl4route2023mao}, and ILRoute \cite{ILRoute2023feng}. Those methods typically utilize learning-based methods to learn the routing strategies/patterns from workers' massive historical behaviors.
    \item \textbf{(Only) time prediction}. Models in this category focus on directly predicting the arrival time of workers without explicitly modeling the route selection process, including  DeepETA \cite{wu2019deepeta}, OFCT \cite{zhu2020order}, CNN-Model \cite{de2021end}, MetaSTP \cite{ruan2022service}, IGT \cite{zhou2023inductive}, DGM-DTE \cite{zhang2023dual}, GSL-QR \cite{zhang2023delivery}.
    \item \textbf{Route and time prediction}. Intuitively, the arrival time of a worker is influenced by his route selection. On the other hand, route selection can also correlate with the arrival time of finished tasks. Therefore, methods in this line learn the joint prediction of route and time, aiming to boost each task's performance by leveraging their mutual correlation, including FDNET \cite{gao2021deep}, RankETPA \cite{wen2023enough}, I$^{2}$RTP \cite{qiang2023i2rtp}, and M$^2$G4RTP \cite{cai2023m2g4rtp}.
\end{itemize}

\subsection{Model Architecture}

\par Model architecture is also an important perspective for classifying different models, including sequence-based models and graph-based models.

\subsubsection{Sequence-based Models}
\par
\begin{figure}[htbp]
    \centering
    \includegraphics[width = 0.75 \columnwidth]{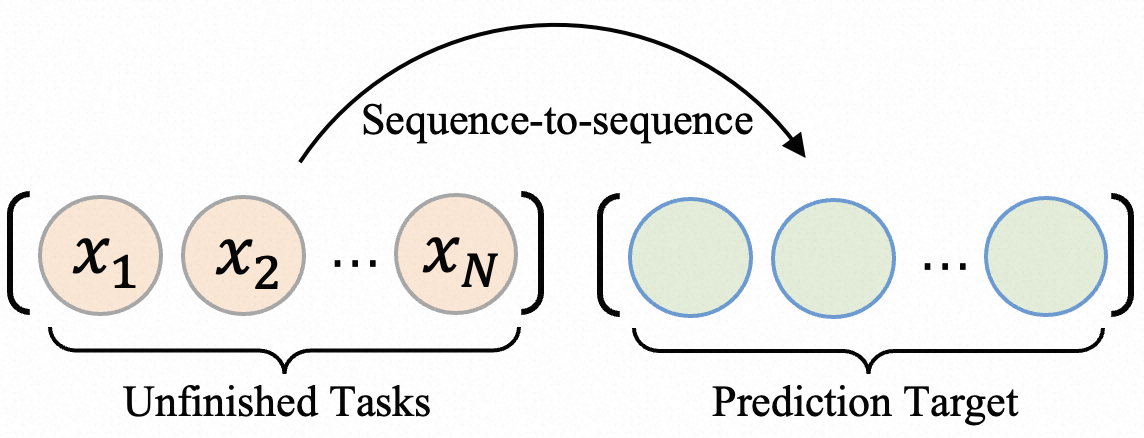}
    \caption{Illustration of sequence-based architecture.}
    \label{fig:seq2seq}
\end{figure}

  As shown in Figure~\ref{fig:seq2seq}, sequence-based models consider the input (i.e., the unfinished tasks) as a sequence, and utilize sequence-to-sequence architecture for solving the related task. These models usually resort to LSTM or Transformer as the encoder to read the input sequence. And use the Pointer-like \cite{Vinyals2015Pointer} decoder to output the desired prediction target. Here we first briefly introduce the two commonly used encoders (LSTM and Transformer), then elaborate on the Pointer-like decoder.

\par \noindent \textbf{LSTM Encoder.} Long Short-Term Memory (LSTM) \cite{lstm1997Hochreiter} is a type of recurrent neural network (RNN) that addresses the vanishing gradient problem by introducing memory cells with self-connected recurrent units. LSTMs are designed to model sequential data and have been widely used in various tasks such as speech recognition \cite{passricha2019hybrid, graves2013hybrid}, natural language processing \cite{yao2018improved, lavanya2021deep}, and time series prediction \cite{hua2019deep, karevan2020transductive, zheng2022hybrid}. The key feature of LSTM is its ability to capture long-term dependencies by utilizing a gating mechanism that controls the information flow within the network. This mechanism involves three main gates: the input gate, the forget gate, and the output gate. LSTM can be formulated in Equation~\ref{equ:lstm}. To ease the presentation, variables are defined locally with a little notion confusion to previous definitions.
\begin{equation}
\begin{aligned}
f_t &= \sigma({\mat W_f} \cdot [{\vec h}_{t-1}, {\vec x}_t] + {\vec b}_f) \\
i_t &= \sigma({\mat W_i} \cdot [{\vec h}_{t-1}, {\vec x}_t] + {\vec b}_i) \\
o_t &= \sigma({\mat W_o} \cdot [{\vec h}_{t-1}, {\vec x}_t] + {\vec b}_o) \\
\tilde{\vec c}_t &= \tanh({\mat W_c} \cdot [{\vec h}_{t-1}, {\vec x}_t] + {\vec b}_c) \\
{\vec c}_t &= f_t \cdot {\vec c}_{t-1} + i_t \cdot \tilde{\vec c}_t \\
{\vec h}_t &= o_t \cdot \tanh({\vec c}_t),
\end{aligned}
\label{equ:lstm}
\end{equation}
where \({\vec x}_t\) is the input at time step \(t\), \({\vec h}_t\) is the hidden state at time step \(t\), \(\vec c_t\) is the cell state at time step \(t\), \({\mat W_f, \mat W_i, \mat W_o, \mat W_c}\) are weight matrices, \(\vec b_f, \vec b_i, \vec b_o, \vec b_c\) are bias vectors, and \(\sigma\) denotes the sigmoid function.

\par \noindent \textbf{Transformer Encoder.} Transformer \cite{vaswani2017attention}  encoder is a key component of the Transformer architecture \cite{vaswani2017attention}, which has revolutionized the field of natural language processing. Unlike traditional recurrent neural networks (RNNs) or convolutional neural networks (CNNs) \cite{luo2017deep}, Transformer encoder relies solely on self-attention mechanisms to capture dependencies between different words or tokens in a sequence. This self-attention mechanism allows the Transformer to efficiently model pairwise long-range dependencies, making it particularly effective for tasks involving sequential data. Here in the RTP problem, each task can be viewed as an item in the sequence. Specifically, the Transformer encoder consists of several transformer blocks, with each equipped with two layers (i) the Multi-Head self-Attention (MHA) layer and (ii) Feed-Forward Network (FFN) layer. MHA layer is formulated in Equation~\ref{eq:self_attention}.
\begin{equation}
    \small
    \begin{aligned}
    \text{Attention}(\mat Q, \mat K, \mat V) &= \text{softmax}\left(\frac{{\mat Q \mat K}^T}{\sqrt{d_k}}\right){\mat V} \\
    \text{MultiHead}(\mat Q, \mat K, \mat V) &= \text{concat}(\text{head}_1, \ldots, \text{head}_h){\mat W_O} \\
    \text{head}_i &= \text{Attention}({\mat Q}{\mat W}_{\mat Qi}, {\mat K}{\mat W}_{\mat Ki}, {\mat V}{\mat W}_{\mat Vi}).
    \end{aligned}
    \label{eq:self_attention}
\end{equation}
Here, $\mat Q$, $\mat K$, and $\mat V$ represent the query, key, and value matrices, respectively. In the self-attention mechanism, all of them are projected from the same input (in our case, the embedding matrix $\mat E$ of all unfinished tasks). $d_k$ denotes the dimensionality of the key vectors. The MHA layer computes the attention weights between the query and key vectors, and the resulting weighted values are then nonlinearly transformed by the FFN layer and concatenated to produce the final output. More complex mutual correlations are captured by stacking multiple transformer blocks.

\par \noindent \textbf{PointerNet Decoder}. Pointer Networks  (PointerNet) \cite{Vinyals2015Pointer} is a type of neural network developed to tackle sequence-to-sequence tasks with varying output lengths. Unlike traditional sequence-to-sequence models, which rely on discrete symbol generation, Pointer Networks learn to output pointers to positions in the input sequence. This makes them particularly useful in the route and time prediction problem, where the output length is not fixed. The core idea of Pointer Networks is the use of an attention mechanism to dynamically select an element from the input sequence as the output at each decoding step. Specifically, PointerNet adopts an RNN such as LSTM as its backbone network. And the equations for the attention mechanism in Pointer Networks are as follows:

\begin{equation}
    \begin{aligned}
        u_{i}^{j} &= {\vec v}^T \tanh({\mat W}_1 {\vec e}_i + {\mat W}_2 {\vec h}_j) \\
        \alpha_{i}^{j} &= \text{softmax}(u_{i}^{j}) \\
        o_j &= \sum_{i=1}^{N}\alpha_{i}^{j}{\vec e}_i.
    \end{aligned}
    \label{eq:pointernet}
\end{equation}
Here, ${\vec e}_i$ represents the encoded representation of the $i$-th item in the input sequence, ${\vec h}_j$ denotes the hidden state of the decoder at step $j$, and $u_{i}^{j}$ represents the compatibility score between the $i$-th input element and the $j$-th decoder state. The attention mechanism calculates the attention weights $\alpha_{i}^{j}$ by applying a softmax function to the compatibility scores.  $o_j$ is the weighted sum of the input sequence $\{\vec e_1, \dots, {\vec e}_{N}\}$ using the attention weights. The core idea of PointerNet is that 
the attention weight $\alpha_{i}^{j}$ can be further regarded as the output probability of item $i$ in the decoding step $j$, which can be regarded as pointers directing to the input.  
Benefit from the above properties,  Pointer Networks have shown promising results in various domains, including routing problems \cite{Kool2019AttentionLT,stohy2021hybrid}, graph optimization \cite{ma2019combinatorial,yang2022graph}, and ranking problem \cite{mottini2017deep, bello2018seq2slate}. 

\subsubsection{Graph-based Models}
\par
\begin{figure}[htbp]
    \centering
    \includegraphics[width = 0.8 \columnwidth]{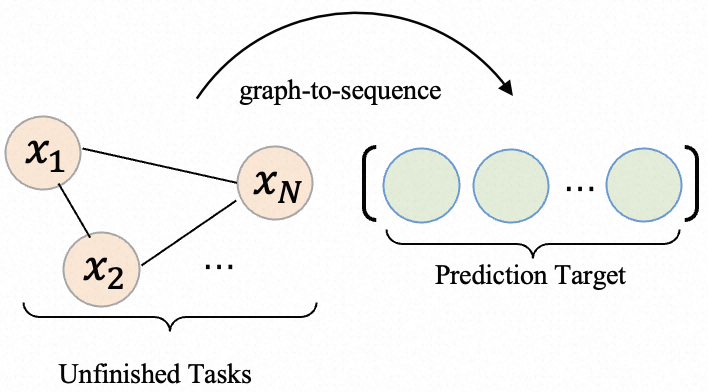}
    \vspace{-0.5em}
    \caption{Illustration of graph-based architecture.}
    \vspace{-0.5em}
    \label{fig:graph2seq}
\end{figure}

 To effectively capture the spatial correlations between different tasks, graph-based models are introduced. As shown in Figure~\ref{fig:graph2seq}, graph-based models consider the input as a graph, and utilize graph-to-sequence architecture for solving the RTP problem.

\par \noindent \textbf{GNN Encoder.}  Graph Neural Network (GNN) \cite{chebnet-2016}, has emerged as the dominant tool for graph data mining \cite{GCN-kipf2017,survey-GNN-2020}. Due to their powerful ability in modeling pair-wise correlation, GNNs \cite{wu2020graph,abadal2021computing,survey-GNN-2020, xu2018powerful} have been widely used in different domains such as node classification \cite{abu2020n, 10.1145/3397271.3401308, zhou2019meta}, graph classification \cite{jin2020certified,cangea2018towards, gao2021higher} and link prediction \cite{rossi2021knowledge, NEURIPS2018_53f0d7c5, zhang2018link}. 

\par Given a graph $G = (\mat{X}, \mat{A})$ with $N$ nodes, where $\mat{X} \in \mathbb{R}^{N \times d_x}$ is the node feature matrix, $d_x$ is the feature dimension. $\mat{A} \in \mathbb{R}^{N \times N}$ is the adjacent matrix of the graph. A general formulation \cite{survey-GNN-2020} of graph neural network can be described as:
\begin{equation}\label{eq:gconv}
    \mat{H} =\sigma\left(\Phi\left(\mat{A},\mat{X}\right) \mat{W}\right),
\end{equation}
where $\mat{W} \in \mathbb{R}^{d_x \times d_x}$ denotes a trainable parameter and $\sigma$ denotes the activation function. $\Phi\left(\mat{A},\mat{X}\right)$ is a function (or a rule) that depicts how neighbors' features are aggregated into the target node. From the above formulation, we can see that one of the core tasks for GNNs is to develop an effective aggregation function $\Phi(\cdot)$. Generally, methods can be classified into two streams: 
\par 1) Spectral-based aggregation, where the graph spectral filter is adopted to smooth the input nodes features. For example, ChebNet~\cite{chebnet-2016} uses the Chebyshev polynomial to optimize the Laplacians decomposition, which reduces the computational complexity. Following ChebNet, the most popular vanilla GNN~\cite{GCN-kipf2017} defines a symmetric normalized summation function as \[\Phi\left(\mat{A},\mat{H}^{l-1}\right) = \mat{\tilde A}\mat{H}^{l-1},\] where \[\mat{\tilde A}=\mat{D}^{-\frac{1}{2}}(\mat{A}+\mat{I})\mat{D}^{-\frac{1}{2}} \in \mathbb{R}^{N\times N}\] is a normalized adjacent matrix. $\mat{I}$ is the identity matrix and $\mat{D}$ is the diagonal degree matrix with $\mat{D}_{ii}=\sum_{j}(\mat{A}+\mat{I})_{ij}$. 

\par 2) Spatial-based aggregation. Unlike spectral-based GNNs, which operate in the spectral domain by exploiting the eigenvectors of the graph Laplacian, spatial-based GNNs focus on aggregating features directly from the spatial domain. These models incorporate spatial convolutions or pooling operations that aggregate and propagate information based on the spatial proximity of nodes in the graph. For example, GraphSAGE~\cite{ hamilton2017inductive} samples a fixed number of neighbors for each node and updates the features, reducing the memory complexity. GAT~\cite{velivckovic2018graph} uses the attention mechanism to adjust the weight of all neighbor nodes. Compared with spectral-based GNNs, spatial-based GNNs have got more attention because of their flexibility in designing the aggregation function.

\par \noindent \textbf{Graph-based Decoder.} Graph-based decoder typically adopts the same architecture as PointerNet, where the attention mechanism is utilized to select candidate nodes and output the route recurrently. Furthermore, the Graph-based decoder tends to incorporate graph information as prior knowledge in the decoding process.  Doing so can improve the accuracy and robustness of predicted routes. For example, Graph2Route \cite{graph2route} constrains the candidate nodes into the neighbors of the outputted node in the last decoding step. ILRoute \cite{ILRoute2023feng} aims to select a node which is the $k$-nearset nodes of the previous outputted node.

\subsection{Learning Paradigm}
\par
\begin{figure}[htbp]
    \centering
    \includegraphics[width = 0.8 \columnwidth]{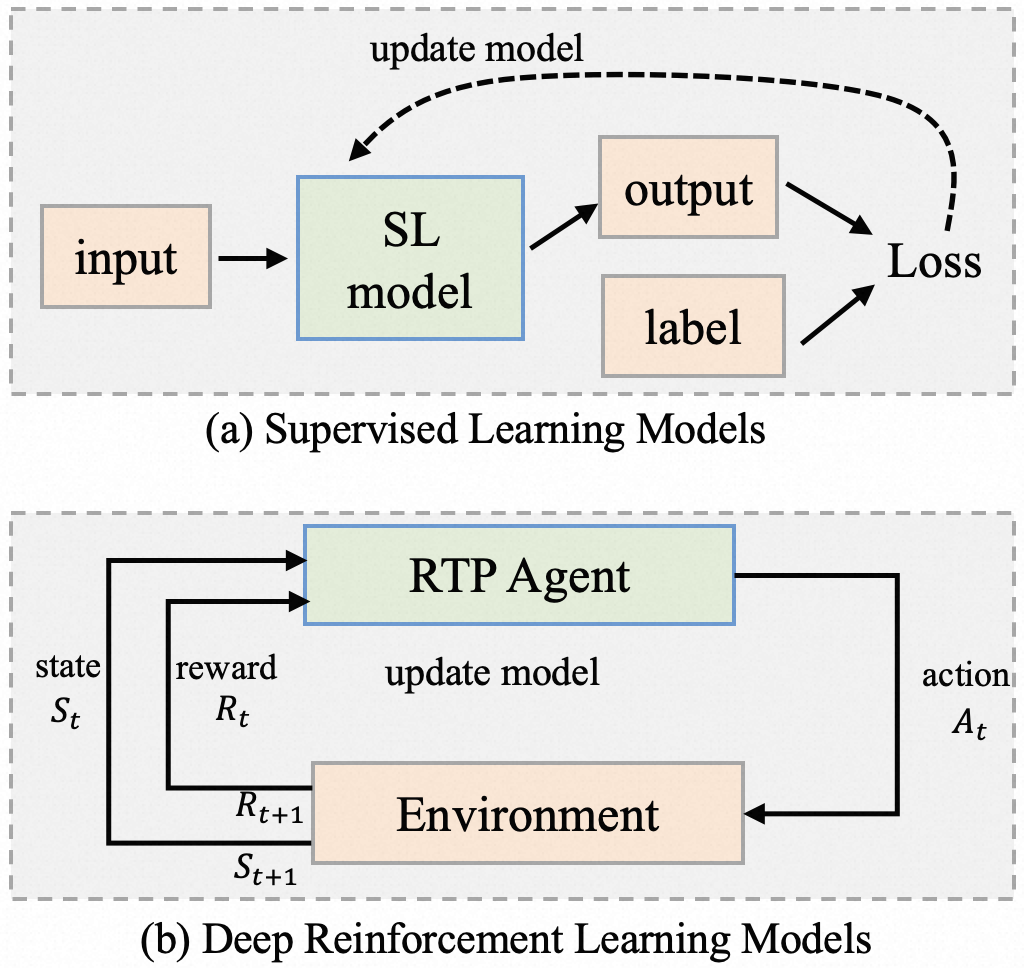}
    \vspace{-0.5em}
    \caption{Illustration of SL-based and DRL-based models.}
    \vspace{-0.5em}
    \label{fig:sl_drl}
\end{figure}

 Models for RTP can also be classified by learning paradigm, which contains two categories, including Supervised Learning (SL-based) and Deep Reinforcement Learning (DRL-based). We illustrate the overall architecture of the two types in Figure~\ref{fig:sl_drl}.

\par SL-based models learn from labeled training data to make predictions for unseen instances. Here in our case, they learn from the data constructed by workers' massive historical behaviors. This technique is widely used in tasks such as image recognition \cite{yang2020tensor}, natural language processing \cite{traylor2019classifying, wang2022fine}, and recommendation systems \cite{chaturvedi2017recommender, wu2022survey}.
% cite: image recognition: kim2006intelligent,

\par On the other hand, DRL-based models combine the principles of deep learning and reinforcement learning to enable the model to learn through interaction with an environment. It involves an agent that takes actions in an environment, receives feedback in the form of rewards/penalties, and learns to optimize its behavior over time. Deep reinforcement learning has achieved remarkable successes in complex tasks such as game playing \cite{lample2017playing, goldwaser2020deep, ye2020towards}, robotics \cite{kober2013reinforcement, zhang2015towards}, and autonomous driving \cite{sallab2017deep,wang2018deep,kiran2021deep}, showcasing its ability to learn directly from raw sensory data and acquire sophisticated decision-making abilities. In the RTP problem, one can consider the model as a route/time prediction agent, to mimic the route selection action of the worker. To this end, DRL methods can be applied to effectively improve the performance of route and time prediction. 
% cite:game playing,  
% cite: autonomous driving:,
\par In summary, we propose to classify the current RTP models from three perspectives, including the task type, model architecture and learning paradigm in this section. Since the RTP problem is a rising topic in the research community, there is still a lack of models for the topic. Therefore, in the next section, we will dive into each category and introduce the details of models to help a comprehensive understanding of each model's motivation and model design.

\begin{table*}[!t]
    \caption{The summary for neural architectures of existing route and time prediction models.}
    \vspace{-1em}
    \small
    \renewcommand\arraystretch{1.35}
    \begin{center}
        \resizebox{1\linewidth}{!}{
            \begin{tabular}{c|c|c|c|c|c|c|c}
                % \hline
                \toprule
                \textbf{Model} & \textbf{Year} & \textbf{Input Information} & \textbf{Problem} & \textbf{Model} & \textbf{Learning paradigm} & \textbf{Task Encoder} & \textbf{Decoder} \\ \midrule
                OSquare\cite{e_le_me} & 2019 & Unfinished Tasks & RP & Sequence-based & SL & LightGBM & LightGBM \\ \hline
                DeepRoute\cite{wen2021package} & 2021 & Unfinished Tasks & RP & Sequence-based & SL & Transformer & Pointer \\ \hline
                DeepRoute+\cite{DeepRoute+} & 2021 & Finish \& Unfinished Tasks, Workers Features & RP & Sequence-based & SL & Transformer & Pointer \\ \hline

                CP-Route\cite{wen2023modeling} & 2023 & Unfinished Tasks & RP & Sequence-based & SL & STC-STP & STC-STP \\ \hline
                
                Graph2Route\cite{graph2route} & 2022 & Finish \& Unfinished Tasks, Workers Features & RP & Graph-based & SL & Dynamic GNN & Graph-based Pointer \\ \hline
                DRL4Route\cite{drl4route2023mao} & 2023 & Unfinished Tasks & RP & Sequence-based & DRL & Transformer & Pointer \\ \hline
                ILRoute\cite{ILRoute2023feng} & 2023 & Unfinished Tasks & RP & Graph-based & DRL & GNN & Pointer \\ \hline

                DeepETA\cite{wu2019deepeta} & 2019 & Finished tasks, Unfinished Tasks & TP & Sequence-based & SL & BiLSTM & MLP \\ \hline
                OFCT\cite{zhu2020order} & 2022 &  Unfinished Tasks & TP & Sequence-based & SL & MLP & MLP \\ \hline
                MetaSTP \cite{ruan2022service} & 2022 &  Unfinished Tasks & TP & Sequence-based & SL & Transformer & MLP \\ \hline
                CNN-model \cite{de2021end} & 2021 &  Unfinished Tasks & TP & Sequence-based & SL & CNN & MLP \\ \bottomrule

                IGT \cite{zhou2023inductive} & 2023 &  Unfinished Tasks & TP & Graph-based & SL & Hete-GCN & Transformer \\ \bottomrule
                DGM-DTE \cite{zhang2023dual} & 2023 &  Unfinished Tasks & TP & Graph-based & SL & Dual GNN & Multi-task, MLP \\ \bottomrule

                GSL-QR \cite{zhang2023delivery} & 2023 &  Unfinished Tasks & TP & Graph-based & SL & GSL, GNN & Attention \\ \bottomrule

                RankETPA \cite{wen2023enough} & 2023 &  Unfinished Tasks & RTP & Sequence-based & SL & Transformer & Pointer \\ \hline
                I$^2$RTP \cite{qiang2023i2rtp} & 2023 & Unfinished Tasks & RTP & Sequence-based & SL & Transformer & Pointer \\ \hline
                FDNET \cite{gao2021deep} & 2021 & Unfinished Tasks, Workers Features & RTP & Sequence-based & SL & DeepFM & Pointer \\ \hline
                M$^2$G4RTP \cite{cai2023m2g4rtp} & 2023 & Unfinished Tasks, Workers Features & RTP & Graph-based & SL & GAT-e &  Multi-task, Pointer \\ \hline

            \end{tabular}
        }
    \end{center}
    \label{tab:all_models}
   
\end{table*}

\section{Service Route Prediction} \label{sec:route_prediction}
\par To facilitate the following sections, we first propose a framework that summarizes the current models. It follows the encoder-decoder structure where we identify four key components in it, including the input construction, task encoder, route decoder, and masked loss.

\par \noindent \textbf{Input.} This component constructs the model's input (i.e., a problem instance) according to the finished tasks and unfinished tasks that contain both spatial and temporal information. A problem instance ${s}_t$ can be represented by a task sequence or task graph, which depends on different methods.

\par \noindent \textbf{Task Encoder.} The Task Encoder  learns the unfinished task representations ${\mat E}_{t} \in \mathbb{R}^{n \times d_e}$ at time $t$ by taking the problem instance $s_t$ as input. Abstractly, we write

\begin{equation}
    {\mat E}_t = { \textbf{TaskEncoder}} (s_t).
    \label{eq:framework_DynGNN}
\end{equation}
It is designed to capture each task's spatial features (e.g., the distance between the task and the worker) and temporal features (e.g., the remaining required time), as well as model the  ST-correlations between different tasks. Here we only list the unfinished task embedding as the output, as it is the core input of the route decoder component in the next step. One can add additional output in this step accordingly.
%, as we will introduce in the following section.

\par \noindent \textbf{Route Decoder.} The decoder computes the predicted route ${\hat {\bm \pi}}_{t:}$ based on the embedding matrix ${\mat E}_t$ outputted by the encoder, equipped with the task decoding module and the service-dependent mask mechanism. The service-dependent mask mechanism is designed to meet the route constraints $\mathcal C$ during the decoding process, specifically, masking unfeasible tasks at each decoding step. Note that the mask mechanism is service-dependent since different types of service can have different route constraints (as we have introduced in Section~\ref{sec:preliminaries}). And the task decoding mechanism is utilized to select a candidate (an unfinished task) at each decoding step. Some works also consider the worker $w$'s personalized feature $\vec x_w$ into the decoding process, thus the overall  route decoder is formulated as:
\begin{equation}
    {\hat {\bm \pi}}_{t:} = {\textbf{RouteDecoder}} ({\mat E}_t, {\vec x_w}),
\end{equation}

% Moreover, different from route planning, where the routing objective function is the same for all workers, e.g., minimizing the total travel distance. In route prediction, different workers have different goals (or objective functions) in route selection, and a worker $w$'s preference in route choices is often unobservable to the service provider. That is, although a worker's objective function is clear to the worker himself, it is unknown to the service provider. Therefore, it is necessary to model the worker in the decoder, and to learn personalized decision preferences from massive historical data, formulated as
% \begin{equation}
%     {\hat {\bm \pi}}_{t:} = {\textbf{PersonalizedRoute-Dec}} ({\mathbf H}_t,  w),
% \end{equation}

\par \noindent \textbf{Masked Loss.} As we mentioned before,  in some service scenarios, there can be new tasks dispatched to the worker at any time. In that case, the new coming task at $t^{\prime}$ can change the worker's previous decisions at $t$, making observations after $t^{\prime}$ inaccurate \cite{wen2021package, gao2021deep} as training label for the sample at time $t$. Therefore, most existing works choose the route observations between $t$ and $t^{\prime}$ (i.e., ${\bm \pi}_{t:{t^{'}}}$) as the label information when training the model, where $t^{\prime}$ is the dispatch time of the first coming task after $t$. In other words, the observation after $t^{\prime}$ is masked when calculating loss. Therefore, we call it ``Masked loss'', which can be formulated as:

\begin{equation}
    \mathcal{L} = {\textbf{MaskedLoss}} ({\hat {\bm \pi}}_{t:}, {\bm \pi}_{t:{t^{'}}}).
\end{equation}

\par To conclude, the Input component represents the problem instance with abundant spatial-temporal information. The Task Encoder is supposed to fully capture the spatial-temporal relationship between different tasks. The Route Decoder component decodes the future route based on the encoded task embedding with or without the worker's personalized information. And the Masked Loss component is designed to eliminate the effects of future new coming tasks on the loss calculation of the current sample. Different models have different customization on the input, task encoder, and route decoder. The following part will introduce how those components are implemented in those models.

% \begin{table*}[]
%         \caption{The summary for neural architectures of existing route prediction models.}
% 	\small
%         \begin{center}
%         \resizebox{1 \linewidth}{!}{
%         \begin{tabular}{c|c|c|c|c|c|c|c}
%         \hline
%         \textbf{Model} & \textbf{Input Information} & \textbf{Model} & \textbf{Task Encoder} & \textbf{Route Decoder} & \textbf{Loss} & \textbf{Scenario} & \textbf{Year} \\ \hline
%         OSquare\cite{e_le_me}  & Unfinished Tasks  & Sequence-based & LightGBM & LightGBM  &  &  & 2019 \\ \hline
%         DeepRoute\cite{wen2021package}   & Unfinished Tasks & Sequence-based & Transformer & Pointer &  &  & 2021 \\ \hline
%         DeepRoute+\cite{DeepRoute+}     & Finish \& Unfinished Tasks, Workers Features  & Sequence-based & Transformer & Pointer &  &  & 2021 \\ \hline
%         FDNET\cite{gao2021deep} & Unfinished Tasks, Workers Features & Sequence-based & BiLSTM & BiLSTM &  &  & 2021 \\ \hline
%         Graph2Route\cite{graph2route}    & Finish \& Unfinished Tasks, Workers Features   & Graph-based & Dynamic GNN & Graph-based + Pointer &  &  & 2022 \\ \hline
%         I$^2$RTP       & Unfinished Tasks  & Sequence-based  & Transformer & Pointer &  &  & 2023 \\ \hline
%         M$^2$G4RTP     & Unfinished Tasks, Workers Features & Graph-based & \blue{GAT-e}  & Pointer &  &  & 2023 \\ \hline
%         \end{tabular}}
%         \end{center}
% \end{table*}

\subsection{Sequence-based SL Models}
\par Sequence-based supervised learning models construct the first research line among all methods for the RTP problem. Methods in this research line include OSquare \cite{e_le_me}, DeepRoute \cite{wen2021package}, DeepRoute+ \cite{DeepRoute+}.

\par  \noindent \textbf{Osquare \cite{e_le_me}.} Osquare is a machine-learning method that treats route prediction as a next-location prediction problem. Algorithm~\ref{algo:OSquare} shows the implementation details of OSquare. It utilizes a point-wise ranking method that trains a traditional machine learning model (i.e., LightGBM \cite{DBLP:conf/nips/KeMFWCMYL17}) to output the probability of all candidates (i.e., $\hat {\vec y}$) at each step, and the one with the maximum probability as the next task. At last, the whole route is generated recurrently.  Overall, the Input component of OSquare constructs a sequence of features. Both the Task Encoder and RouteDecoder are composed of LightGBM.
\floatname{algorithm}{Algorithm}  
\renewcommand{\algorithmicrequire}{\textbf{Input:}}  
\renewcommand{\algorithmicensure}{\textbf{Output:}}  
\begin{algorithm}  
	\caption{OSquare.}  
	\begin{algorithmic}[1] %每行显示行号  
% 		\Require features of unfinished tasks of worker $w$ at time t $X_t=\{{\mathbf x}_t1, {\mathbf x}_t2 ... {\mathbf x}_tn\}$; features of current status $s=(l, t)$, where $l$ is courier's current location $l$ and $t$ is current time; max number of unfinished tasks $\rm max$; padding vector $z$.
    	\Require features of unfinished tasks of worker $w$ at time $t$ ${\mat X_t}=\{{\vec x}_{1}, {\bm x}_{2} ... ,{\bm x}_{n}\}$; max number of unfinished tasks $\rm max$; padding vector $\vec z$.
		
		\Ensure predicted pick-up route $\pihat$%{\bm{\hat \pi}}
		\State $\pihat \gets [ ]$;
	
		\For{$j = 1, ..., n$}
			\For {$m \in \bm{\hat \pi}$}   
			    \State ${\mat X_t}[m] = {\vec z}; \text{  //pad tasks outputted before}$
			\EndFor
	
		\State ${\mat X}_t^{\prime} \gets {\rm{concatenate}}({\vec x}_{1}, ..., {\vec x}_{n}, {\vec z}_{n+1}, ..., {\vec z}_{{\rm max}})$;
		
		\State ${\hat {\vec y}} = {{\rm LightGBM}}({\mat X}_t^{\prime})$;
		\State ${\hat \pi_j} \gets \mathop {{\rm{argmax}}}_k {\rm{ }}{\hat {\vec y}_k}, ~{\rm where} ~k \in \{1, \dots, n\} {\rm{~and~}}  k \not \in {\pihat} $;
		\State $\pihat \gets \pihat + [\hat \pi_j]$;
        \EndFor	
	\end{algorithmic}  
	\label{algo:OSquare}
\end{algorithm}

\par \noindent \textbf{DeepRoute \cite{wen2021package}.} DeepRoute is the first deep neural network proposed for the package pick-up route prediction problem. Unlike OSquare, it is a list-wise model that ranks all unfinished tasks at once. Specifically, the implementation of different components in DeepRoute are: 
\begin{itemize}[leftmargin=*]
    \item \textbf{Sequence Input}. The input of DeepRoute is a sequence that contains features of unfinished tasks introduced in Section~\ref{sec:preliminaries} that may affect a courier's routing decision. An unpicked-up package represents a task in DeepRoute.
    \item \textbf{Transformer-based Task Encoder}. DeepRoute adopts the Transformer Encoder to model the spatial-temporal correlation between different tasks, no matter the distance of two tasks in the given sequence.
    \item \textbf{Pointer-based Route Decoder}. PointerNet decoder is utilized as the backbone network to output the tasks step by step. Moreover, one route constraint is that no duplicated outputted is allowed in the route prediction problem. To meet the constraint, DeepRoute adopts the mask mechanism that masks the outputted tasks before. In that case, the compatibility score at decoding step $j$ in Equation~\ref{eq:pointernet} can be rewritten as:
    \begin{equation}
        \small
        u_i^j = \left\{ {\begin{array}{*{20}{l}}
        	{{{\vec v}^T}\tanh ({\mat W_1}{\vec e_i} + {\mat W_2}{\vec h_j})}&{{\rm{ if }}{\kern 1pt} {\kern 1pt} i \ne {\pi _{j'}}\quad \forall j' < j}\\
        	{ - \infty }&{{\rm{ otherwise,}}}
        	\end{array}} \right.	
    \label{eq:deeproute_decoder}
    \end{equation}
    where $\vec e_i$ is the encoded embedding of task $i$, and $\vec h_j$ is the hidden state of the decoding step $j$. And the output probability of task $i$ at deciding step $j$, i.e., $y_i^j$ is calculated by the softmax of the compatibility score, formulated as $y_i^j={\rm softmax}(u_i^j)$.
    
\end{itemize}

\par \noindent \textbf{DeepRoute+ \cite{DeepRoute+}}. DeepRoute+ is an improved version of DeepRoute, which models workers' personalized features:
\begin{itemize}[leftmargin=*]
    \item \textbf{Sequence Input}. DeepRoute mainly focuses on modeling the spatial-temporal factors that influence the worker's routing decision. Compared with DeepRoute, DeepRoute+ also models the worker's personalized preference by taking their features $\vec x_w$ and the latest finished tasks ${\mathcal{O}_t^{f}}$ as input. 
    \item \textbf{Preference-aware Task Encoder}. A worker decision preference module is designed to identify which factors have an important impact on the worker's decision under the current situation. It learns a mapping function to map the worker's individual features (including total working days and average pick-up number) and his latest finished task (encoded by BiLSTM) sequence to a decision preference vector $\vec p$ of the worker. Then the decision vector is merged into the Transformer encoder by updating the task embedding $\vec e_i$ using Hardamard product: ${\vec e_i} = {\vec p} \odot {\vec e_i}$. 
    \item \textbf{Pointer-based Route Decoder}. Like DeepRoute, DeepRoute+ also adopts the pointer-based route decoder to output the predicted route.
\end{itemize}

\par \noindent \textbf{CP-Route \cite{wen2023modeling}}. CP-Route aims to model the personal information of workers by mining their spatial transfer patterns.
\begin{itemize}[leftmargin=*]
    \item \textbf{Sequence Input}. CP-Route takes two sequences as input: i) the unfinished tasks, ii) their corresponding AOI ID.
    \item \textbf{STC-STP Encoder.} it contains two encoder blocks: an STP-aware location embedding block, which aims to incorporate workers' Spatial Transfer Patterns (STP) into the location embedding; and a correlation-aware constraints embedding block, which aims to incorporate the Spatial-Temporal Correlations (STC) into the task embeddings.
    \item \textbf{STC-STP Decoder.} A mixed-distribution-based decoder is designed to simultaneously consider the influence of STC and STP on couriers' final decisions.
\end{itemize}

\subsection{Sequence-based DRL Models} 

\par
\begin{figure}[hbtp]
		\centering
		\includegraphics[width=1 \columnwidth]{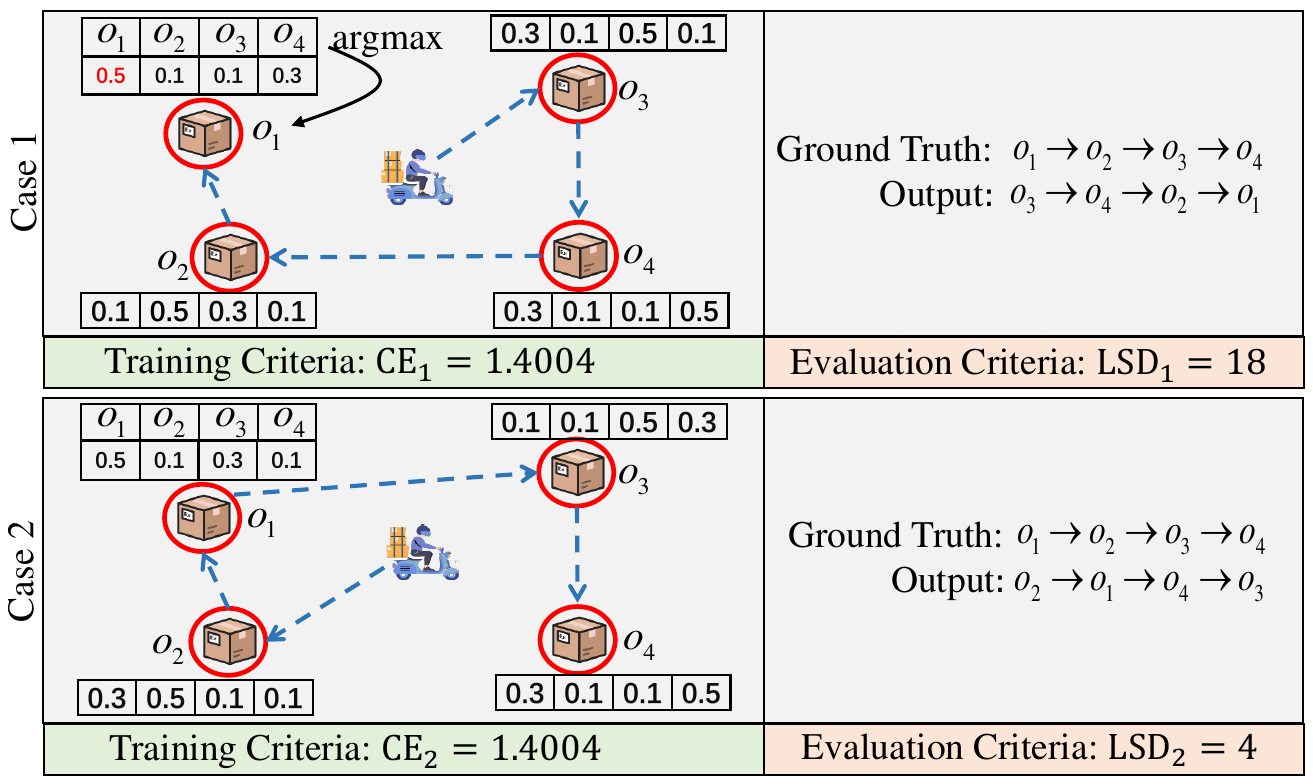}
		% \captionsetup{font={small}}
		\caption{Illustration of mismatch between the training and test objectives. The vector is the output probability corresponding to the task.}
	\label{fig:drl_motivation}
\end{figure}

 The above sequence-based SL models suffer from the limitation where the training criteria is not the same as the test one. Specifically, those methods consider the task selection at each step as a classification problem, train the model using the Cross-Entropy (CE) as the loss function, while evaluating the model using other measurements, such as LSD \cite{graph2route} and KRC \cite{graph2route}. Thus leading to a mismatch between the training and test objectives. Taking Figure~\ref{fig:drl_motivation} as an example, despite producing the same value on the training criteria (i.e., CE), the two cases exhibit quite different results on the test criteria (i.e., LSD). This disparity limits the potential of a ``well-trained" model to deliver more favorable performance in terms of the test criteria,  which considerately trims down their performance when applied in real-world systems. Consequently,  Sequence-based DRL models are proposed to distinguish these two cases during the training process.

\par Currently, models in this category only contain DRL4Route. Here we first introduce how the route prediction is formulated from the RL perspective. Then elaborate on the model architecture of DRL4Route.

\subsubsection{Formulation from the RL perspective}

\par Route prediction can be considered as a sequential decision-making process, where each task on the route is outputted step by step based on previous decisions. It can be modeled as a discrete finite-horizon discounted Markov Decision Process (MDP) \cite{sutton1998introduction}, in which a route prediction agent interacts with the environment and makes decisions over $T$ time steps. Formally, MDP is denoted by $M = (S, A, P, R, s_0, \gamma, T)$, where $S$ is the set of states, $\emph{A}$ is the set of actions, $P: S \times A \times S \rightarrow {\mathbb R}_{+}$ is the transition probability, $\emph{R}: \emph{S} \times A \rightarrow {\mathbb R}$ is the reward function, $s_0: \emph{S} \rightarrow {\mathbb R}_{+}$ is the initial state distribution, $\gamma \in \left[0, 1\right]$ is a discount factor, and $T$ is the total time steps determined by the number of unfinished tasks (in our case $T$ equals the number of unfinished tasks $n$). We introduce the details of the agent, state, action, reward and state transition probability in the following part. 

\par \noindent \textbf{Route Prediction Agent.} The route prediction agent selects a task from the unfinished task candidates step by step, which can be implemented based on the aforementioned sequence-based SL models.

\par \noindent \textbf{State.} The state $s_j \in \mathcal{S}$  represents the environment's condition at the $j$-th ($j \in \{1, \dots, n\}$) decoding step. It encompasses the relevant information that enables the agent to make decisions at each decoding step. The state is formulated as $s_j = ({\mat E}, \mathcal{C}, {\vec h_j}, \pihat_{1:j-1})$, where ${\mat E}$ is the encoded embedding matrix of unfinished task, $\mathcal{C}$ is the route constraints, ${\vec h_j}$ is the hidden state of the $j$-th step, and $\pihat_{1:j-1}$ denotes the route generated by the agent up to the $j$-th decoding step.

\par \noindent \textbf{Action.} An action $a_j \in \mathcal{A}_j$ refers to the selection of a task $\pi_j$ based on the current task candidates and states. A joint action $(a_1, \cdots, a_n) \in \mathcal{A}=\mathcal{A}_1 \times \cdots \times \mathcal{A}_n$ forms a predicted route. The action space $\mathcal{A}_j$ specifies the available task candidates that the agent can choose from at the $j$-th step. It changes during the decoding process because of the route constraints.

\par \noindent \textbf{Reward.} The reward is defined based on the test criteria to align the training and test objectives. Here different rewards can be designed according to different test objectives. Equation~\ref{eq:drl4route_reward} shows the reward definition of DRL4Route, whose core idea is giving rewards to actions that are close to the label route based on LSD:
\begin{equation}
\small
r_j = \left\{ 
    {\begin{array}{*{20}{l}}
        -\mathrm{LSD}{({n^{\prime}} + 1, j)}&  {\hat \pi}_j \notin {\bm \pi}, j \leq {n^{\prime}}, ({\rm case~1})\\
    	0 &  {\hat \pi}_j \notin {\bm \pi}, j > {n^{\prime}}, ({\rm case~2})\\
    	  - \mathrm{LSD}{(Y_{{\hat \pi}}({{\hat \pi}_j}) + 1, j)} &  {\hat \pi}_j \in {\bm \pi}, j \neq {\hat \pi}_j, ({\rm case~3}) \\
        \overline{R} & {\hat \pi}_j \in {\bm \pi}, j = {\hat \pi}_j, ({\rm case~4})
    \end{array}} \right.
    \label{eq:drl4route_reward}
\end{equation}
where  $\overline{R}$ is a hyper-parameter to control the scale of the cumulative reward. $Y_{\hat \pi}(\pihat_j)$ is the order of task $\pihat_j$ in the predicted route. And $n^{\prime}$ is the number of tasks in the label route $\bm \pi$.

\par \noindent \textbf{State Transition Probability.} The state transition probability $P(s_{j+1}|s_j, a_j): \mathcal{S} \times \mathcal{A} \times \mathcal{S} \rightarrow \mathbb{R}_{+}$ represents the likelihood of transition from state $s_j$ to $s_{j+1}$ when action $a_j$ is taken at state $s_j$. In DRL4Route, the environment is considered deterministic, meaning that the resulting state $s_{j+1}$ after taking action $a_j$ from state $s_j$ is predetermined and certain.

\par \noindent \textbf{Definition 9: RL-based RP Problem.} Given a state $s_j$ at time $j$, the route prediction agent generates an action by the current policy $\pi_\theta$ parameterized by $\theta$, then receives the task-specific reward $r_j$ from the environment. The training goal of RL-based methods is to learn the best parameter $\theta^{*}$ of the route prediction agent that can maximize the expected cumulative reward, formulated as:
\begin{equation}
    \theta^{*} = {\arg\max}_{\theta}{\mathbb{E}_{\pi_{\theta}}\left[\mathop\sum\limits_{j=1}^{n}{\gamma^j}r_j\right]},
\end{equation}
where $\gamma$ is the discount factor that controls the tradeoffs between the importance of immediate and future rewards. 

\subsubsection{DRL4Route Architecture}

\begin{figure}[htbp]%hbtp
		\centering
		\includegraphics[width=1          
            \columnwidth]{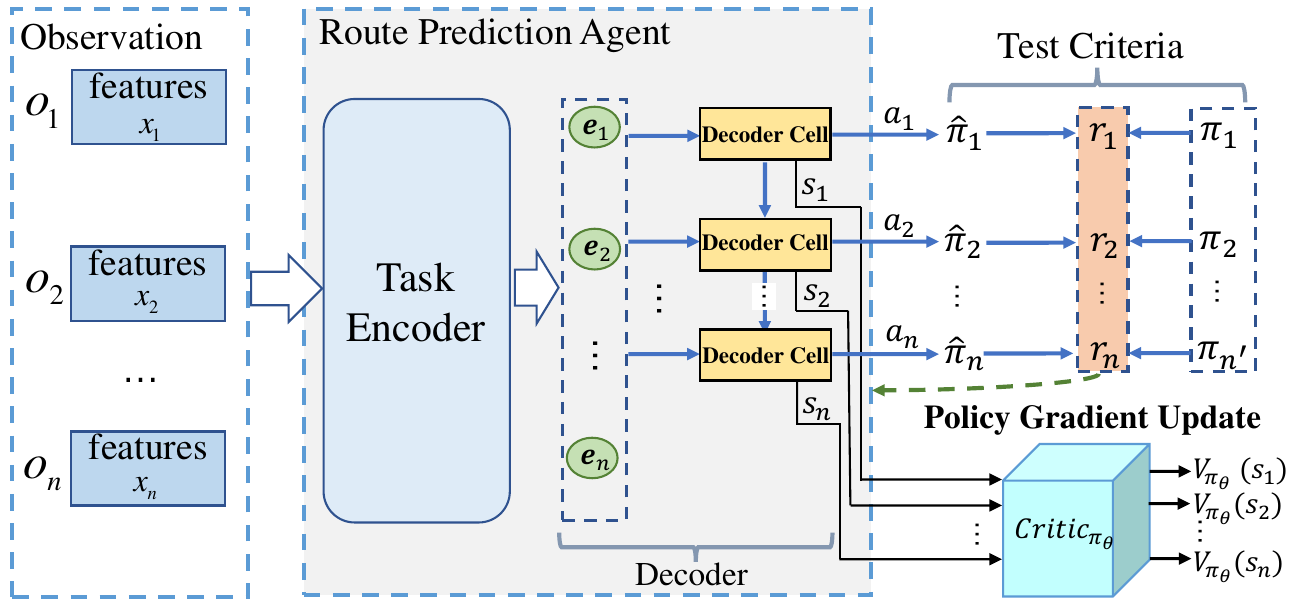}
		\caption{{DRL4Route Framework \cite{drl4route2023mao}.}}
		\label{fig:drl4route_framework}	
\end{figure}

\par The overall architecture of DRL4Route is depicted in Figure~\ref{fig:drl4route_framework}. It adopts an Actor-Critic architecture, which reduces the variance of the policy gradient estimates by providing a reward right after each action. The ``Actor'' is the route prediction agent which updates its policy under the guidance of the ``Critic''. And the ``Critic'' estimates two functions, namely i) the state-value function $V$ to evaluate the value for a state and ii) the state-action function $Q$ to evaluate the benefits of taking a certain action under a certain state. Given a policy $\pi_{\theta}$, the two functions are defined as:
\begin{equation}
    Q_{\pi_{\theta}}(s_j, a_j) = \mathbb{E}_{\pi_{\theta}}\left[r({\hat \pi}_j,\cdots,{\hat \pi}_n)|s=s_j,a=a_j\right],
\end{equation}
\begin{equation}
    V_{\pi_{\theta}}(s_j) = \mathbb{E}_{a_j\sim {\pi_{\theta}}(s_j)}\left[Q_{\pi_{\theta}}(s_j, a = a_j)\right].
\end{equation}
Furthermore, we can use $V$ function to estimate $Q$ function as shown in Equation~\ref{eq:Q_V_relation}. Doing so can reduce the number of estimated functions, thus reducing the risk of estimation error. 
\begin{equation}
    Q_{\pi_{\theta}}(s_j, a_j) =  \mathbb{E}[r_j + \gamma * V_{\pi_{\theta}}(s_{j+1})].
    \label{eq:Q_V_relation}
\end{equation}
Some previous efforts \cite{2020Keneshloo} find that removing the exception calculation can significantly accelerate the training process while achieving promising results, formulated as:
$Q_{\pi_{\theta}}(s_j, a_j) =  r_j + \gamma * V_{\pi_{\theta}}(s_{j+1})$. 
\vspace{0.3em}

\par Based on the above formulation, advantage function $A_{\pi_{\theta}}$ is defined as subtracting the value function $V$ from the $Q$-function, which is used to reflect the relative superiority of each action and update the model parameters: 
\begin{equation}
    \begin{aligned}
        A_{\pi_{\theta}}(s_j, a_j) &= Q_{\pi_{\theta}}(s_j, a_j) - V_{\pi_\theta}(s_j) \\
          &  \approx r_j + \gamma V_{\pi_{\theta}}({s_{j+1}}) - V_{\pi_{\theta}}(s_j).  \\
    \end{aligned}
    \label{eq:eq_advantage}
\end{equation}
\par \noindent \textbf{Training.} Overall, the actor first accumulates thousands of samples by current policy. Based on the generated samples, the critic learns and updates the $V$ function, which is further used to calculate the advantage approximation function $A_{\pi_{\theta}}(s, a)$. At last, the actor is trained by the following loss function:
\begin{equation}
{\mathcal L_{\rm actor}}={\frac{1}{K}} \mathop\sum\limits_{k=1}^{K}\sum_{j = 1}^{n}{A_{\pi_\theta}}(s_{k, j}, a_{k, j}){\rm log}{{\pi}_{\theta}(a_{k, j}|s_{k, j})},
\label{eq:loss_actor}
\end{equation}
where $K$ is the total number of samples. And the critic is trained via a robust regression loss \cite{girshick2015fast}, which is less sensitive to outliers than $L_2$ loss:
\begin{equation}
{\mathcal L_{\rm critic}}={\frac{1}{K}} \mathop\sum\limits_{k=1}^{K}\sum_{j=1}^{n}\mathrm{smooth}{L_1}({\hat V}(s_{k, j}) - r({\hat \pi}_{k,j},\cdots,{\hat \pi}_{k,n})),
\label{eq:loss_critic}
\end{equation}
in which $\mathrm{smooth}{L_1}$ is defined as
\begin{equation}
\mathrm{smooth}{L_1}(x) = 
\left\{ 
{\begin{array}{*{20}{l}}
    0.5x^2&  |x|<1,\\
	|x|-0.5& \mathrm{otherwise}.\\
	\end{array}} \right.
\label{l1_loss}
\end{equation}

\subsection{Graph-based SL Models} 

\par The sequential nature of the above sequence-based methods limits their ability to fully encode the spatial-temporal correlations between different tasks. To overcome the limitations of the sequence-based encoders, graph-based algorithms model a problem instance from the graph perspective and take full advantage of the node/edge features and graph structure of all tasks. A representative method is Graph2Route. We first introduce the problem formulation from the graph perspective, and then we introduce the details of Graph2Route.

\subsubsection{Formulation from Graph Perspective}

\par In real scenarios, service tasks are essentially located in different geographic areas. The spatial relationship of those tasks can be naturally described as a graph. Therefore,  some works formulate the route prediction task from the graph perspective, which can better represent the intrinsic relationship in a problem instance.

\par \noindent \textbf{Definition 10: Input ST-Graph.}  A problem instance of worker $w$ at time $t$ can be defined on a spatial-temporal graph (ST-graph) $\mathcal{G}_t^{w}=(\mathcal{V}_t, \mathcal{E}_t, \mat{X}_t^{v}, \mat{X}_t^{e})$, where $\mathcal{V}_t=\{v_1, \dots, v_{m+n}\}={\mathcal{O}_t^{f}} \cup {\mathcal{O}_t^{u}}$ contains both $m$ finished tasks and $n$ unfinished tasks, with each node corresponds to a task of the worker. 
${\mathcal{E}_t}=\{(i,j)~|~v_i,v_j \in \mathcal{V}_t\}$ is the set of edges. To ease the presentation, let $\overline{n}=m+n$. ${\mat X}_t^{v} \in \mathbb{R}^{\overline{n} \times d_v} $ and ${\mat X}_t^{e}  \in \mathbb{R}^{\overline{n} \times \overline{n} \times d_e}$ are the node and edge features respectively, where $d_v$ and $d_e$ are the node feature dimension and edge feature dimension, respectively. Both of them contain the spatial-temporal features of different tasks and can be constructed by service-specific settings. 

%Moreover, let $\mathbf{x} \in \mathbb{R}^{d_v} $ be the features of a task $v$. And $x^{FT}$ is one of the features, denoting the finish time of the task. At a certain time $t$, there are two types of nodes in graph $\mathcal{G}_t^w$: i) finished nodes $\mathcal{V}_t^F = \{ v~|~v \in {\mathcal{V}_t},{x^{FT}} \le t\}$, and ii) unfinished nodes $\mathcal{V}_t^U = \{ v~|~v \in {\mathcal{V}_t},{x^{FT}} =  - 1\}$. 

% For example, constructing nodes features according to the requirements of a task (e.g., location, promised service time), and constructing edge features according to the spatial-temporal correlation between two tasks (e.g., distance, connectivity). 

\par
\begin{figure}[hbtp]
		\centering
		\includegraphics[width=0.9 \columnwidth]{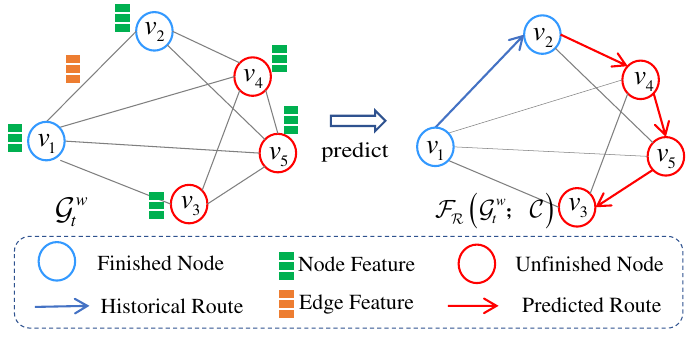}
		%\captionsetup{font={small}}
		\caption{Problem Illustration \cite{graph2route}. In this case,  $\mathcal{O}_t^f=\{v_1, v_2\}$ and $\mathcal{O}_t^u=\{v_3, v_4, v_5\}$, the output of the model $\hat {\bm \pi}=[\hat \pi_1, \hat \pi_2,\hat \pi_3]$ is $[4,5,3]$.  }
		%$\mathcal{V}_t = \{v_1, \dots, v_5\}$
		\label{fig:problem}
  
\end{figure}

 \noindent \textbf{Definition 11: Graph-based RP Problem}. Given the input graph $\mathcal{G}_t^{w}$ of worker $w$ at time $t$, Graph-based RP problem aims to learn a mapping function ${\mathcal F}_{{\mathcal C}}$ to predict the worker's future service route $\hat {\bm \pi}$ of unfinished nodes which can satisfy the given route constraints $\mathcal C$, formulated as:
\begin{equation}
    {\mathcal F}_{R}({\mathcal G}_t^w, {\mathcal C}) = [\hat \pi _1, \hat \pi_2, \cdots, \hat \pi_n], \\
\end{equation}
where $\hat \pi_i$ means that the $i$-th node in the route is node $v_{\pi_i}$. Moreover, ${\hat \pi _i} \in \{ 1, \cdots, n \}~{\rm and}\;{\hat \pi _i} \ne {\hat \pi _j}\;{\rm if}\;i \ne j$. Basically, the graph-based RP problem can be considered a reformulation of the general RP problem from the graph perspective. Figure~\ref{fig:problem} gives an illustration of the problem. And Table~\ref{tab:notation} lists the related notations. 

% and $|\cdot|$ denotes the cardinality of a set

\subsubsection{Graph2Route Architecture}

\par Graph2Route investigates service route prediction from the dynamic graph perspective. Traditional models typically treat problem instances at different time steps as independent, relying solely on the request time for prediction. However, in real-life scenarios, the problem instance for a worker evolves over time, such as changes in graph signals or the arrival of new nodes. This evolution establishes natural connections between different problem instances. In other words, the route arrangement at a specific time $t$ is closely related to previous instances, especially those in proximity. Therefore, Graph2Route defines the decision context at time $t$ as the decision-making environment encompassing the workers' activities several time steps prior. By introducing the decision context (represented as $\Psi$) in the model, Graph2Route can leverage more valuable information throughout the entire service process, resulting in improved accuracy in route prediction.

\begin{itemize}[leftmargin=*]
    \item \textbf{ST-Graph Input}. The input contains graphs from several consecutive time steps. For each graph, the node features ${\mat X}_t^{v}$ are essentially the tasks features introduced in Sectioin~\ref{sec:preliminaries}. The edge set ${\mathcal E}_t$ and edge feature ${\mat X}_t^{e}$ are defined according to the distance of two tasks from the spatial (i.e., coordinates) or temporal (i.e., the required time window) perspective. One important feature is the $k$-nearest neighbors. 
    
    \item \textbf{DynGNN Task Encoder.} To capture the decision context, a dynamic spatial-temporal graph neural network (DynGNN) is developed that models the evolving relationship between consecutive problem instances. In Equation~\ref{eq:framework_DynGNN}, the encoder first computes the node embeddings ${\mat E}_{t} \in \mathbb{R}^{{\overline n} \times d_h}$ and edge embeddings ${\mat Z}_{t} \in \mathbb{R}^{{\overline n} \times {\overline n} \times d_h}$ by a GNN (i.e., the spatial-correlation encoding, Spatial-CE) to leverage their spatial interactions. Then it updates the node embeddings efficiently based on the prior ones (i.e., the decision context $\Psi$) by temporal-correlation encoding (Temporal-CE), which is implemented by an RNN (i.e., GRU \cite{pan2020spatio}) architecture. The dominant advantage of such a way is fully utilizing spatial-temporal features and considering the decision context.

    % It computes the node embeddings ${\mat E}_{t} \in \mathbb{R}^{{\overline n} \times d_h}$ and edge embeddings ${\mat Z}_{t} \in \mathbb{R}^{{\overline n} \times {\overline n} \times d_h}$ at time $t$ by a GNN, while capturing the spatial-temporal relationship between different nodes, including the intra-relationship between unfinished nodes and the inter-relationship between finished and unfinished nodes.

    \begin{equation}
        \begin{aligned}
             {\mat E}_t & = { \text{DynGNN-Enc}} (\mathcal{G}_t^{w}, \Psi) \\
             & = {\text{Temporal-CE}(\text{Spatial-CE}(\mathcal{G}_t^{w}), {\mat E}_{t-1}))}. \\
        \end{aligned}
        \label{eq:framework_DynGNN}
    \end{equation}

    \item \textbf{Graph-based Personalized Route Decoder.} A graph-based decoder is designed to filter extremely unreasonable solutions. The decoder computes the predicted route ${\hat {\bm \pi}}_{t:}$ by a recurrent attention mechanism which selects a node from the graph based on the embedding matrix ${\mat E}_t$. At each decoding step, it only considers the $k$-nearset neighbors of the current node as candidates for the next step. Moreover, the worker's features are also incorporated into the attention mechanism to achieve more Personalized route prediction.

    %Moreover, different from route planning, where the routing objective function is the same for all workers, e.g., minimizing the total travel distance. In route prediction, different workers have different goals (or objective functions) in route selection, and a worker $w$'s preference in route choices is often unobservable to the service provider. That is, although a worker's objective function is clear to the worker himself, it is unknown to the service provider. Therefore, it is necessary to model the worker in the decoder, and to learn the personalized decision preference from massive historical data, formulated as

    % \begin{equation}
    %     {\hat {\bm \pi}}_{t:} = {\textbf{PersonalizedRoute-Dec}} ({\mathbf H}_t, {\mathbf Z}_t, w),
    % \end{equation}
 
\end{itemize}

%\par To conclude, the ST-Graph Input component represents the problem instance by a graph with abundant spatial-temporal information. The DynGNN Encoder component is supposed to fully capture the relationship between nodes and the evolution of the decision context. The Personalized Route Decoder component is supposed to decode the future route by considering the worker's personalized information. 

\subsection{Graph-based DRL Models}
% \par Introduce the ILRoute. \mxw{introduce IRL2route}
\par The graph-based DRL methods combines the advantage of the graph and DRL for route prediction. The method in the class includes ILRoute, which integrates graph neural networks into DRL frameworks to extract the multi-source and heterogeneous features in the workers' decision-making process and unveil workers' routing strategies. ILRoute learns workers' routing strategies by imitation learning and leverages the workers' real route to provide the expert policy. Here we first introduce how route prediction is formulated as MDP in the imitation learning framework, then we elaborate on the architecture of ILRoute.

\subsubsection{Formulation from the imitation learning}

\par In imitation learning, route prediction is formulated using MDP to maximize the accuracy of route prediction. In the MDP, the route prediction agent takes action $a_j$ at the $j$-th step based on the state defined as $s_j=({\vec x_w}, \mathcal{O}_j^{u}, \mathcal{O}_j^{f}, {\mat X}_j^{v}, {\mat X}_j^{e})$, which contains the worker's personalized features, finished and unfinished task features, node and edge features. After an action $a_j$ is taken, the current state $s_j$ transits to the next state $s_{j+1}$. In the state transition, the worker's personalized features remain unchanged. The task features, context features, and route history will change from $s_j$ to $s_{j+1}$ due to new route node choice and time changes.
\par Notably, the reward function in ILRoute is learned from the real worker's routes instead of being defined in advance. This reward function measures the similarity between the route generated by the route generator and the real workers' routes. And the reward value is calculated by the discriminator.

% \par \noindent \textbf{Route Prediction Agent.} The agent makes the route prediction at the requesting time, aiming to imitate the workers' decisions.
% %At a certain time $t$, the worker $w$ is considered as the agent, which observes the worker's personalized features $\bm x_w$, the finished tasks $\mathcal{O}_t^{f}$, the unfinished tasks $\mathcal{O}_t^{u}$, node features ${\mat X}_t^{v}$, and edge features ${\mat X}_t^{e}$.

% \par \noindent \textbf{State.} For agent $w$, the state at the $j$-th step is defined as $s_j=({\vec x_w}, \mathcal{O}_j^{u}, \mathcal{O}_j^{f}, {\mat X}_j^{v}, {\mat X}_j^{e})$, which contains the worker's personalized features, finished and unfinished task features, node and edge features.

% %order features, context features, and route history.

% \par \noindent \textbf{Action.} The action $a_j$ at the $j$-th step is selecting the next location the worker will move to.

% \par \noindent \textbf{Transition.} This component describes how the current state $s_j$ transits to the next state $s_{j+1}$ after an action $a_j$ is taken. Specifically, the worker's personalized features will remain unchanged. The order features, context features, and route history will change from $s_j$ to $s_{j+1}$ due to new route node choice and time changes.

% \par \noindent \textbf{Reward.} This reward function is to measure how similar the route generated by the graph-based route generator compared with the real worker's routes.

\subsubsection{ILRoute Architecture}
\begin{figure}[htbp]
    \centering
    \includegraphics[width=1 \linewidth]{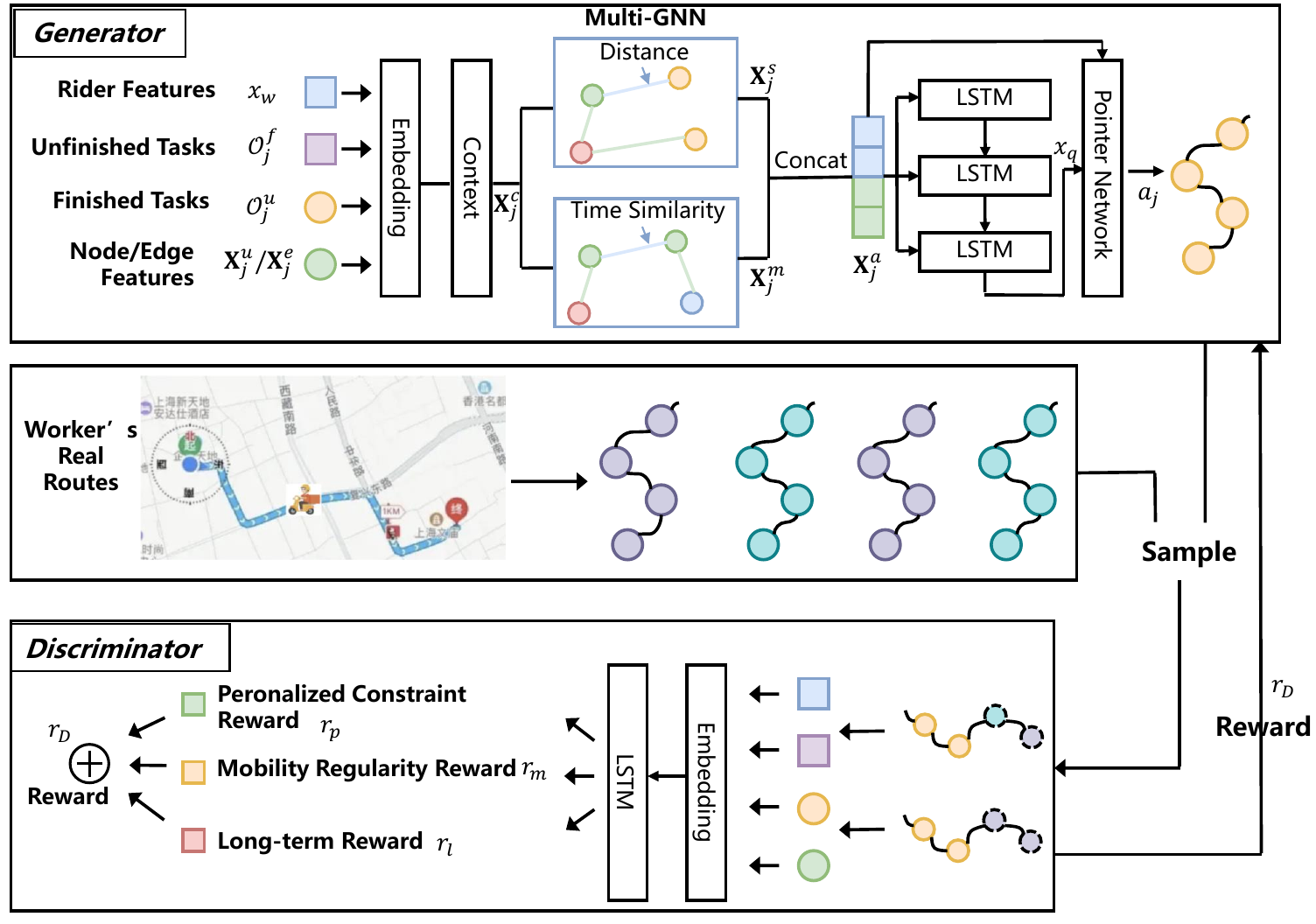}
    \caption{The framework of ILRoute. }
    \label{fig:ILRoute}
    
\end{figure}

\par The overall framework of ILRoute is shown in Figure~\ref{fig:ILRoute}. It is equipped with a graph-based route generator and a sequential discriminator. The graph-based route generator takes the state as input and converts them into the route choice of a worker. The sequential discriminator distinguishes the route generated by the graph-based route generator and returns the reward to the generator to revise its policy.
\par \noindent \textbf{Graph-based Route Generator.} The graph-based route generator denoted as $\pi_{\theta}$ consists of a multi-graph encoder and a PointerNet decoder. The multi-graph encoder is designed to extract multi-source and heterogeneous features as spatial-temporal embeddings and model the complex relationships among features that influence worker's route decisions. Specifically, the spatial-GNN and the temporal-GNN convert vector representation ${\mat X}_j^{c}$ of $j$-th step into distance  ${\mat X}_j^{s}$ and time similarity embedding ${\mat X}_j^{m}$. Then they are concatenated to obtain the hidden embedding ${\mat X}_j^{a}$. The PointerNet decoder is designed to convert the hidden embeddings ${\mat X}_j^{a}$ of the observed route into the next route choice of the worker step by step.

\par \noindent \textbf{Sequential Discriminator.} The sequential discriminator is designed to distinguish the generation quality of the graph-based route generator compared with the real workers' route. It also introduces a mobility regularity-aware constraint to reduce route choice exploration with prior spatial continuity knowledge and a personalized constraint mechanism to enhance the personalization of the worker's route decision-making process. The discriminator takes the whole route as input and utilizes an LSTM and a sigmoid function to convert the input into the long-term reward $r_l$, which is calculated as follows:
    \begin{equation}
        \begin{aligned}
             r_l = {\rm sigmoid}({\rm LSTM}(s_1, s_2, \dots, s_{n})). \\
        \end{aligned}
        \label{eq:long-term reward}
    \end{equation}
%where $T$ is the length of the route.
\par \noindent \textbf{Reward Desigin.} ILRoute introduces a mobility regularity-aware constraint to add an auxiliary reward $r_m$, which assumes that workers will pick up or deliver the nearby tasks first. The calculation of $r_m$ is defined as:
    \begin{equation}
        \begin{aligned}
             r_m =  -\mathop\sum\limits_{j=0}^{n-1}{\rm Distance}(l_j, l_{j+1}), \\
        \end{aligned}
        \label{eq:regularity-aware constraint}
    \end{equation}
where $l_j$ denotes the task's location visited by the worker at the $j$-th step, and $\rm Distance$ denotes the Manhattan distance between two locations.
ILRoute also proposes a personalized constraint mechanism to add the mutual regulation between the generated routes sequences $\hat{\bm \pi}$ and the worker's personalized features ${\vec x_w}$. The mechanism is achieved by maximizing the mutual information $I(\hat{\bm \pi}; \vec x_w)$, which can be calculated as follows:
    \begin{equation}
        \begin{aligned}
             I(\hat{\bm \pi}; {\vec x_w}) & =  H({\vec x_w}) - H({\vec x_w}|\hat{\bm \pi}) \\
             &= H({\vec x_w}) + E_{\hat{\bm \pi}}E_{{\vec x_w}|\hat{\bm \pi}}logp({\vec x_w}|\hat{\bm \pi}) , \\
        \end{aligned}
        \label{eq:mutual_information}
    \end{equation}
where $H$ deontes the entropy value, $E$ denotes the expectation value and $p$ denotes the probability.
\par Without access to the posterior $p({\vec x_w}|\hat{\bm \pi})$, we cannot maximize $I(\hat{\bm \pi}; {\vec x_w})$ directly. Here, $q({\vec x_w})$ is introduced to approximate the true posterior $p({\vec x_w}|\hat{\bm \pi})$:
    \begin{equation}
        \begin{aligned}
             logp({\vec x_w}|\hat{\bm \pi}) = logq({\vec x_w}|\hat{\bm \pi}) + log{\frac{p({\vec x_w}|\hat{\bm \pi})}{q({\vec x_w}|\hat{\bm \pi})}}.
        \end{aligned}
        \label{eq:posterior}
    \end{equation}
Take Equation~\ref{eq:posterior} into Equation~\ref{eq:mutual_information}, it can be observed that $E_{{\vec x_w}|\hat{\bm \pi}}log{\frac{p({\vec x_w}|\hat{\bm \pi})}{q({\vec x_w}|\hat{\bm \pi})}}$ is always larger than $0$. Through the reparameterization trick \cite{kingma2013auto}, the left part of Equation~\ref{eq:mutual_information} can be expressed as follows:
    \begin{equation}
        \begin{aligned}
             I(\hat{\bm \pi}; \vec x_w) &\geq \int p(\vec x_w)logq(\vec x_w|\hat{\bm \pi}) + H(\vec x_w) \\
             &\equiv D_{KL}(p(\vec x_w|\hat{\bm \pi})||q(\vec x_w|\hat{\bm \pi})).
        \end{aligned}
        \label{eq:reparameterization}
    \end{equation}
To this end, maximizing $I(\hat{\bm \pi}; \vec x_w)$ can be achieved by maximizing $D_{KL}(p(\vec x_w|\hat{\bm \pi})||q(\vec x_w|\hat{\bm \pi}))$. Based on this, a personalized constraint reward $r_p = D_{KL}(p(\vec x_w|\hat{\bm \pi})||q(\vec x_w|\hat{\bm \pi}))$ is added to enhance the personalization of the worker's route decision-making process. Therefore, the reward $r_D$ of the discriminator can be obtained as follows:
    \begin{equation}
        \begin{aligned}
             r_D = r_l + {\beta}{r_m} + {\gamma}{r_p},
        \end{aligned}
        \label{eq:reward_of_discriminator}
    \end{equation}
where $\beta$ and $\gamma$ are hyperparameters to control the scale of different rewards.
\par The discriminator is denoted as $D_{\phi}$, which is parameterized by $\phi$ and is optimized based on the following loss function:
    \begin{equation}
        \resizebox{1 \linewidth}{!}{
        $\begin{aligned}
            {\mathcal L_{D}}=-\mathbb{E}_{\pi_{\hat{\bm \pi}}}[logD_{\phi}(\hat{\bm \pi})]-\mathbb{E}_{\pi_{\theta}}[log(1-D_{\phi}(\hat{\bm \pi}))]-\mathbb{E}_{\pi_{\theta}}[logq(\vec x_w|\hat{\bm \pi})],
        \end{aligned}$
        
        }
        \label{eq:reward_of_discriminator}
    \end{equation}
where $\mathbb{E}_{\pi_{\hat{\bm \pi}}}$ represents the expectation with respect to the real workers' routes. In addition, $\mathbb{E}_{\pi_{\theta}}$ represents the expectation with respect to the routes generated by generator $\pi_{\theta}$. 
\par \noindent \textbf{Training.} The generator network with parameter $\pi_{\theta}$ and the discriminator with parameter $D_{\phi}$ are trained together in ILRoute. Firstly, the discriminator is trained by considering the generated route as negative samples while the real-world sequences as positive samples. Then, a batch of rewards is calculated for the generated routes. Finally, the generator is trained by maximizing the expectation of reward via the actor-critic algorithm.

\section{Service Time Prediction} \label{sec:time_prediction}
\par Time prediction models directly predict the arrival time of unfinished tasks, without counting on the route estimation. We first briefly introduce the difference between time prediction in instant delivery and another related popular research topic, i.e., estimated time of arrvial (ETA) prediction in map service. 
\par In map service, ETA prediction especially refers to the travel time estimation given a pair of origin and destination locations. Methods in this topic can be classified into two types: i) path-based methods \cite{wang2018will,de2008traffic,sevlian2010travel,jenelius2013travel,qiu2019nei,han2023ieta, liu2023uncertainty}, whose input requires the path information between the origin and destination. ii) path-free methods \cite{hu2020stochastic,wang2019simple,bertsimas2019travel,li2018multi,lin2023origin}, which do not require the path information.

\par Unlike map services, the ETA problem in instant delivery focuses on predicting the time for each task in a given set of unfinished tasks. This prediction is based on the worker's current status, such as location, and is essentially a multi-destination prediction problem, which makes the problem even more challenging. Since the worker can freely decide the route, which is unknown when making the prediction. In that case, the problem setting in our task is distinctly different from the ETA problem in map services. In this section, we focus on introducing methods for instant delivery rather than for map service.

\subsection{Sequence-based SL Models}
%\par \noindent \textbf{LightGBM.} It is a machine-learning model which is commonly adopted in the industry. It treats the time prediction of all unfinished tasks as a multiple-output regression problem. Thus the input of LightGBM is the features of all tasks, and the output is the estimated arrival time $\tauhat$.

\par \noindent \textbf{DeepETA \cite{wu2019deepeta}.} DeepETA aims to predict the arrival time of couriers for package delivery. It mainly models three important factors: i) the sequence of the latest delivery route, ii) the regularity of the delivery pattern, and iii) the sequence of packages to be delivered.
\begin{itemize}[leftmargin=*]
    \item \textbf{Sequence Input}. The input contains the sequence of the latest route and the set of packages to be delivered.
    \item \textbf{Task Encoder.}  Firstly, a lasted route encoder is developed to capture the spatial-temporal and sequential features of the latest delivery route, using a combination of spatial encoding and recurrent cells. Secondly, a frequent pattern encoder is designed with two attention-based layers, which leverage the similarity between the latest route and future destinations to predict the most probable estimated time of arrival (ETA) based on historical frequent and relative delivery routes.
    \item \textbf{Time Decoder.} DeepETA adopts a fully connected layer (MLP) to jointly learn the delivery time and output the results. 
\end{itemize}

\par \noindent \textbf{OFCT \cite{zhu2020order}.} OFCT proposes a deep neural network to predict the Order Fulfillment Cycle Time (OFCT), which refers to the amount of time elapsed between a customer placing an order and he/she receiving the meal. 

\begin{itemize}[leftmargin=*]
    \item \textbf{Sequence Input}. A main contribution of OFCT is the extraction of numerous features that can influence the arrival time, including i) the spatial-temporal information of the task, ii) the supply-and-demand features for describing the supply-and-demand status, and iii)  and couriers' features.
    \item \textbf{Task Encoder.} Equipped with elaborately designed features, OFCT designs a simple model architecture for prediction. As for the encoder, it adopts the fully connected exponential linear units (ELU) \cite{clevert2015fast} and the embedding layer to transform the numerical and categorical features.
    \item \textbf{Time Decoder.} The regression module is a simple MLP with two hidden layers of fully connected ELUs.
    
\end{itemize}

\par \noindent \textbf{CNN-Model \cite{de2021end}.} CNN-Model proposes an end-to-end system capable of parcel delivery time prediction. It studies applying a series of deep Convolutional Neural Networks (CNNs \cite{tahan2022development}) to solve this problem, relying solely on  the start and end points.

\begin{itemize}[leftmargin=*]
    \item \textbf{Sequence Input.} The input contains the latitude and longitude of the depot and the task destination, the accept time of the task, and weather conditions.
    \item \textbf{CNN-based Task Encoder.} It applied and tested three different convolutional network architectures for learning spatio-temporal features of tasks as well as weather features. The first class of convolutional networks is based on VGG modules \cite{simonyan2014very}, which comprises a number of convolutional layers followed by a pooling layer. The second class of convolutional networks is ResNet \cite{he2016deep}, which helps to mitigate the problem of vanishing gradients by a skip connection. The third class of convolutional networks is SE block \cite{hu2018squeeze}, which contains a Squeeze Operator and an Excitation Operator.
    \item \textbf{Time Decoder.} This method utilizes 2 fully connected layers to output the estimated delivery time of each task.

\end{itemize}

\par \noindent \textbf{MetaSTP \cite{ruan2022service}.} MetaSTP  proposes a meta-learning-based neural network model to predict the service time, which is the time spent on delivering tasks at a certain location. 
\begin{itemize}[leftmargin=*]
    \item \textbf{Sequence Input.} The input contains the fine-grained,  aggregated information of undelivered tasks, and context information such as workday and time of day.
    \item \textbf{Task Encoder.} Firstly, a task representation module is developed to extract and embed features of each task, then combines the embeddings with other task features to obtain the fine-grained hidden representation of each task. Secondly, a historical observation encoding module is implemented by self-attention and temporal convolution.  It encodes the correlation between the hidden representation of the query task and tasks with labels in the support set. Finally, a location-wise knowledge fusion module is adopted to further enhance the output of the encoder with the location-prior knowledge.
    \item \textbf{Time Decoder.} MetaSTP utilizes fully connected layers to output the estimated delivery time of each task.
    \item \textbf{Training.} MetaSTP follows the paradigm of model-based meta-learning to extract the meta-knowledge that is globally shared among a set of related learning tasks. In training, each individual learning task $\mathcal{T}$ consists of a support set $\mathcal{D}^s={( {\bm x}_i^s, y_i^s)}^{N_s}_{i=1}$ and a query set $\mathcal{D}^q={({\bm x}_i^q, y_i^q)}^{N_q}_{i=1}$. The inference of each query sample $\bm{x}^q$ is formulated as $\hat{y}^q=f({\bm x}^q, \mathcal{D}^s, \theta)$, where $\theta$ is the meta-knowledge that is globally shared, and $f$ is a neural network parameterized by $\theta$. The inference can also be written as:
      \begin{equation}
        \begin{aligned}
          \hat{y}^q=f_{\theta}(\bm{x}^q, \mathcal{D}^s).
        \end{aligned}
        \label{eq:meta-inference}
    \end{equation}
    To optimize $\theta$, we leverage a set of learning tasks already sampled from a learning task distribution $p(\mathcal{T})$, which is called meta-training tasks $\mathscr{T}_{\rm meta-train}$. The optimal $\theta$ learned from $\mathscr{T}_{\rm meta-train}$ should adapt well to any learning task sampled from $p(\mathcal{T})$ based on Equation~\ref{eq:meta-inference}, which is achieved by optimizing the following meta loss function:
        \begin{equation}
        \begin{aligned}
            \mathscr{L}(\theta) = \mathop\sum\limits_{\mathcal{T}_i \in {\mathscr{T}_{\rm meta-train}}}{\frac{1}{\left| \mathcal{D}^q_i \right |}}{\mathop\sum\limits_{(\bm{x}^q, y^q) \in \mathcal{D}_i^q}}\mathcal{L}(f_{\theta}(\bm{x}^q, \mathcal{D}^s_i), y^q).
        \end{aligned}
        \label{eq:meta-loss}
    \end{equation}
 $\mathcal{D}^s_i$ and $\mathcal{D}^q_i$ are the support and query set of learning task $\mathcal{T}_i$. And $\mathcal{L}$ is the loss function of a learning task.
\end{itemize}

\subsection{Graph-based SL Models}
\par \noindent \textbf{IGT \cite{zhou2023inductive}.} The goal of IGT is similar to OFCT. It aims to predict the time from user payment to package delivery, given the information of retailer, origin, destination, and payment time. IGT proposes an Inductive Graph Transformer (IGT) to address the challenge of inductive inference (i.e., models are required to predict ETA for orders with unseen retailers and addresses) and high-order interaction of order semantic information.

\begin{itemize}[leftmargin=*]
    \item \textbf{Heterogeneous Graph Input.} To fully model the high-order interaction of order semantic information, a heterogeneous graph is constructed, where each element in order (i.e., retailer, origin, destination, and payment time) is represented as a node in the graph. Two nodes are linked if they occur in the same order. IGT further limits the links according to proposed rules (such as the retailer node can only connect to the origin node) to reduce the density of the graph and the computational complexity.
    \item \textbf{Temporal and Heterogeneous GCN Encoder} (Hete-GCN). To model the heterogeneous graph, it first constructs a set of bipartite subgraphs based on the combinations of the node types. Then the graph convolution is performed in each bipartite subgraph. Until now, the node embedding has been injected with information of graph structure and the inter-correlations between different nodes. Based on the learned node embedding, IGT further adopts GRU to analyze temporal correlations on the time-series axis at the node level.  
    \item \textbf{Transformer-based Time Decoder}. In the decoder, the raw features of an order and the embedded order embedding are encapsulated into the same vector and then fed into the Transformer for the estimation.
\end{itemize}

\par \noindent \textbf{DGM-DTE \cite{zhang2023dual}.} The studied problem of DGM-DTE is the same as IGT.  DGM-DTE targets the challenge of imbalanced delivery time estimations and proposes a dual graph multitask framework.

\begin{itemize}[leftmargin=*]
    \item \textbf{Multi-graph input.} Given the retailer, origin, destination, and payment time information of orders, DGM-DTE constructs three graphs named spatial, temporal, and merchant relation graphs. The spatial relation graph composed of OD pairs (i.e., sending and receiving addresses). The temporal graph represents  the periodicity of payment time in weeks and days. And the merchant graph represents the similarity (defined by historical order) between merchants.
    \item \textbf{Dual Graph-based Encoder.} To tackle the imbalanced delivery time estimations, it first classifies the input into two classes: head and tail data according to the delivery time distribution. Then, it designs two graph-based representation brunches where GCN and GAT are employed. One learns high-shot data representation in head data, and another re-weights the representations of tail data according to kernel density estimation \cite{he2014nonlinear} of labels.
    \item \textbf{Multitask Decoder}. The order representations learned from the dual graph module are then aggregated and fed into a DNN predictor for time estimation. Dosing so the model can focus on both high-shot regional data and rare labeled data. Moreover,  DGM-DTE actually adopts a multitask learning framework that predicts delivery time from two-view, i.e., 1) the classification of the head or tail data and 2) the imbalanced data regression.
    
\end{itemize}

\par \noindent \textbf{GSL-QR \cite{zhang2023delivery}.}  GSL-QR improves the model performance, by learning the optimal graph structure and graph embeddings guided by the downstream ETA task.

%The studied problem of GSL-QR  is the same as IGT, aiming to predict the time from user payment to package delivery.  

\begin{itemize}[leftmargin=*]
    \item \textbf{Spatial and Temporal Relation Graph Input.} Given the sending and receiving addresses, payment time, and merchant information of orders, GSL-QR constructs spatial and temporal relation graphs. The spatial relation graph is a similarity relation graph among OD pairs (sending and receiving addresses) built upon their spatial attributes. The edge represents the similarity relation between two OD pairs, and the edge weight is the similarity score. The temporal relation graph is a similarity relation graph among the payment time of orders. Each node is a tuple indicating the payment time of an order placed on the day of the week and the hour of the day.
    \item \textbf{GSL and GNN Encoder.} GSL-QR proposes a Graph Structure Learning (GSL) method for simultaneously learning the optimal relation graph structure and potential node embedding for ETA prediction. It uses a metric learning-based graph learner that first obtains node embeddings from the initial spatial relation graph, then reconstructs the adjacency matrix based on the pairwise similarity of node embeddings. GAT is utilized in both learning function and node embedding generation. For the temporal relationship graph, GSL-QR uses a similar method, while GCN is utilized in the learning function and node embedding generation.
  
    \item \textbf{Attention-based Decoder}. GSL-QR propose a multi-attribute adaptive attention aggregation for dynamically measuring the contributions of the spatial, temporal, and context attribute. A DNN is used for final regression prediction. GSL-QR argues that not only the accuracy of ETA prediction, but also the order fulfillment rate should be considered. To strike a balance between them, GSL-QR employs quantile regression to find an optimal point.
    
\end{itemize}

\section{Joint Route and Time  Prediction} \label{sec:route_and_time_prediction}

\par Models within this category aim to predict both the future route and the arrival time of a worker. The rationale behind this is that the arrival time and route are often highly correlated with each other. Existing works mainly cover sequence-based SL models and graph-based SL models.
\subsection{Sequence-based SL Models}

Three methods are included in this class, including RankETPA \cite{wen2023enough}, FDNET \cite{gao2021deep}, and I$^2$Route \cite{qiang2023i2rtp}.

\par \noindent \textbf{RankETPA \cite{wen2023enough}.} RankETPA  develops a two-step model for package pick-up route and time prediction. It first predicts the future pick-up route, then feed the pick-up route as the input for the time prediction.
\begin{itemize}[leftmargin=*]
    \item \textbf{Sequence Input.} The input contains the features of unfinished tasks as described in the preliminary.
    \item \textbf{Task Encoder.} RankETPA has a route predictor and a time predictor. The encoder of the route predictor is LSTM, which reads the input step by step. While the encoder of the time predictor is Transformer, which encodes the sequential information of the predicted route and the correlation between different tasks.
    \item \textbf{Route\&Time Decoder.} The decoder of the route predictor is PointerNet, and the decoder of the time predictor is also Transformer. The route predictor first estimates the future service route, which is converted into the positional encoding and fed into the time decoder.
\end{itemize}

\par \noindent \textbf{FDNET \cite{gao2021deep}.} FDNET  is a deep learning method to tackle the food delivery route and time prediction task. 
\begin{itemize}[leftmargin=*]
    \item \textbf{Sequence Input.} The input of FDNET contains the features of unfinished tasks and workers' features.
    \item \textbf{Task Encoder.} FDNET has two modules: RP (route prediction) and TP (time prediction). Both of them treat the input as a sequence and share the same LSTM as the encoder. Moreover, DeepFM \cite{guo2017deepfm} is adopted to learn the interactions between different features.
    \item \textbf{Route\&Time Decoder.} The RP module predicts the probability of each feasible location the worker will visit next and generates the complete delivery route. A model based on RNN and attention is designed to depict the behavior decision process of drivers based on features affecting drivers' behaviors. The TP module predicts the travel time duration between two adjacent locations (from leaving the previous location to arriving at the next one) in the route. A Wide and Deep model \cite{cheng2016wide} is designed to predict the time duration based on the built context, worker and spatiotemporal features. For each step, input locations of the TP module are produced by the RP module, and the result of the TP module will be used to update features for predicting the worker's future behaviors. %Meanwhile, the worker's personalized information is introduced to mine their delivery habits and abilities.
\end{itemize}
% It generates the arrival time at each location combined with the earliest pickup time of each order.

\par \noindent \textbf{I$^2$Route \cite{qiang2023i2rtp}}. I$^2$Route is proposed for package delivery route and time prediction. It is the first model that explores the inter- and intra-community routing patterns of workers.
\begin{itemize}[leftmargin=*]
    \item \textbf{Sequence Input.} I$^2$Route aims to explore the case where only limited information is available in the system. Its input features only contain the latitude, longitude, and community id of the package. 
    \item \textbf{Inter- and Intra-Community Task Encoder.}  I$^2$Route has two modules: i) the inter-community module with LSTM learns how workers transfer in different communities; ii) The intra-learning module pays attention to the trajectory of a worker inside a community and the time duration between consecutive tasks. It adopts the Transformer encoder to capture the correlation between different tasks.
    \item \textbf{Two-level Route\&Time Decoder.} I$^2$Route explicitly model the inter-community and intra-community transition behavior by two separate PointerNet-based nets. Especially for prediction inside a community, it designs a residual-based block to integrally predict the next task and the time duration between two consecutive tasks inside a community. 
\end{itemize}

\subsection{Graph-based SL Methods}

\begin{figure}[htp]
    \centerline{\includegraphics[width=1 \linewidth]{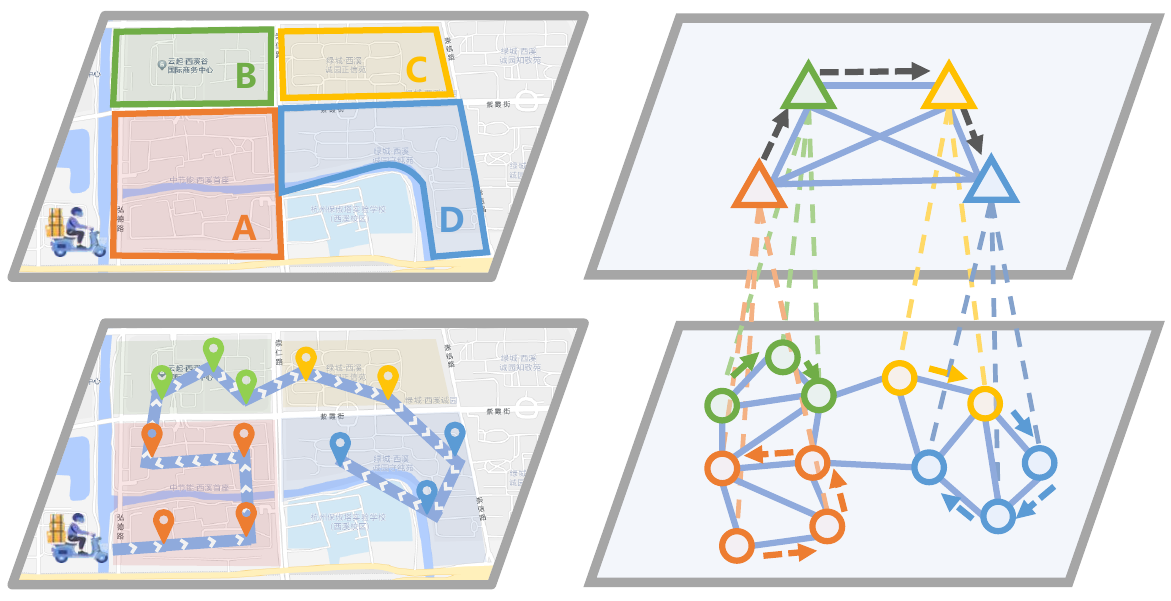}}
    \caption{Modoling the problem from the multi-level graph perspective.}
    \label{fig:multi-level-graph}
\end{figure}
\par \noindent \textbf{M$^2$G4RTP \cite{cai2023m2g4rtp}}. M$^2$G4RTP proposes a multi-level and multi-task graph model for joint route and time prediction in the logistics platform. 
\begin{itemize}[leftmargin=*]
    \item \textbf{Multi-level Graph Input.} To model both AOI-level and location-level transfer patterns, a multi-level graph is constructed where AOIs and locations are treated as nodes. We illustrate the multi-level graph in Figure~\ref{fig:multi-level-graph}.
  
    \item \textbf{Multi-level Graph Task Encoder.} M$^2$G4RTP develops a multi-level graph encoder, which is equipped with GAT-e (graph attention network \cite{velivckovic2018graph} with edge feature accounted) encoding module for modeling workers' high-level transfer mode between AOIs and low-level transfer mode between locations.
    \item \textbf{Multi-task and Multi-level Route\&Time Decoder.} A multi-task and multi-level decoder completes both the route and time prediction in a multi-task manner for the location- and AOI-level, respectively. It is composed of an AOI-level decoder to rank all AOIs and provide guidance for the location-level decoder. In the location-level decoder, tasks inside an AOI are ranked, and the arrival time is outputted. Besides, during the training process, the route prediction (classification task) and the time prediction (regression task) are trained together in a multi-task manner. Classification and regression are two heterogeneous tasks, and the loss function values are in different scales. To tackle this, M$^2$G4RTP uses the weight assignment technic based on homoscedastic uncertainty \cite{kendall2018multi} to balance these tasks in the training process.
  
\end{itemize}

%\par 4) Future research directions: We identify the limitations of current works and discuss the potential future research directions in route prediction, to inspire innovative ideas and promote growth within the domain.

%In Section 4, we offer a quantitative comparison of the discussed algorithms, pinpointing prevalent trends and patterns in contemporary approaches, while also highlighting constraints and potential avenues for future research.

\subsection{Applications} 
\par
\begin{figure}[htbp]
    \begin{minipage}{1\linewidth}	 
        \subfigure[Intelligent Order Sorting Service]{
            \label{application:courier}
            \includegraphics[width=0.48\linewidth]{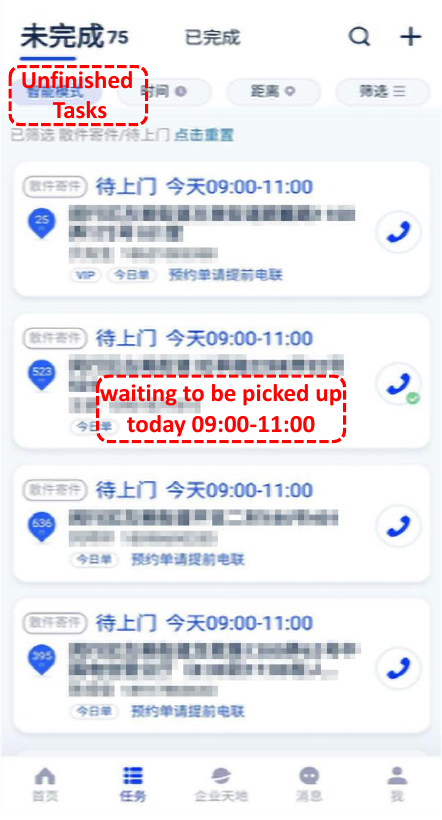}
        }\noindent
        \subfigure[Minute-level ETA Service]{
            \label{application:user}
            \includegraphics[width=0.48\linewidth]{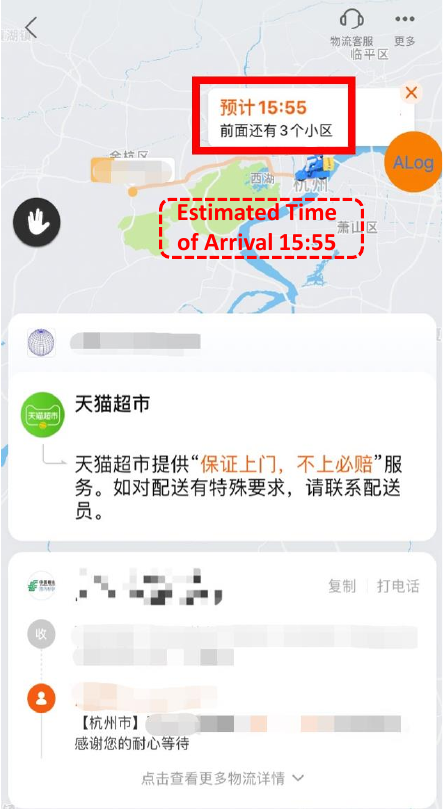}
        }
    \end{minipage}
    
    \caption{Applications on Cainiao APP \cite{cai2023m2g4rtp}.}
    \label{fig:application}
\end{figure}

 To intuitively show how the RTP methods can serve real-world systems, we illustrate two applications from the Cainiao system in Figure~\ref{fig:application}.

% \begin{figure}[htp]
%     \centerline{\includegraphics[width=1 \linewidth]{img/application.pdf}}
%     \caption{Illustration of two real-world applications \cite{wu2023lade}. (a): Route prediction predicts the future pick-up route of a courier. (b): ETA prediction estimates the courier's arrival time for picking up or delivering packages.}
%     \label{fig:application}
% \end{figure}

\par As shown in Figure~\ref{application:courier}, the Intelligent Order Sorting Service is designed to assist couriers during package pick-up. In this setting, pickup tasks are generated randomly, as users can place orders at any time. This requires couriers to frequently update their route plans due to incoming tasks. Route prediction simplifies the courier's job by intelligently sorting orders based on the courier's likely future route. Before this service, the platform could only display all pending orders in either a time-focused or distance-focused manner. This forced couriers to sift through all orders to plan their routes. With intelligent sorting, the order list now aligns better with couriers' work habits, easing the burden of route planning for couriers.

\par As shown in Figure \ref{application:user}, package pick-up is a face-to-face service, requiring customers to be available until the courier arrives. This often leads to increased waiting anxiety for users, making accurate ETA (Estimated Time of Arrival) crucial. The improved route and time prediction now offer users a more reliable and accurate ETA service. Previously, the platform's ETA service gave a broad 2-hour window, allowing the courier to arrive at any point within that period. With the new precise route and time prediction, the platform can now offer minute-level ETAs and inform users about the number of remaining orders before the courier arrives.

\section{Limitation and Future Direction} 
\label{sec:limitation_future_direction}
\par In this section, we first elaborate on the limitations of the current work, then discuss some promising directions for research in this field.

% Limitation: 对于路线预测的方法，1）解码器都是循环解码器，在input数量多的时候效率会比较低。ii）没有考虑到路网信息 iii）开源数据和代码比较少。
\subsection{Limitations}
\par  Though current methods have achieved promising performance, there are still some limitations. 
\par \textbf{Recurrent decoding mechanism.} Firstly, most decoders for route/time prediction adopt the recurrent architecture \cite{wen2021package, graph2route, DeepRoute+, gao2021deep, cai2023m2g4rtp}, where the prediction targets are calculated step by step, and the output of the previous step is usually fed as input for the next step. To this end, the recurrent architecture may encounter the efficiency problem. Especially in real-world scenarios such as last-mile delivery, a work (i.e., courier) can have around (even more than) 50 tasks at the same time, which brings big challenges for the recurrent decoder.
%especially when the number of tasks expands.  

\par \textbf{Lack of modeling the road network.} Secondly, all current models did not take the road network into consideration. Most of them only model the spatial-temporal features regarding the finished or unfinished tasks. Some works also model the additional geographical information, such as the community \cite{qiang2023i2rtp} or AOI \cite{cai2023m2g4rtp}. Nevertheless, all ignore the road network, which is a natural and important spatial information. Ignoring such information can notably compromise the model's efficacy.

\par \textbf{Error accumulation in time prediction.} Thirdly, current solutions for RTP typically utilize the route prediction results as the input for time prediction, such as \cite{qiang2023i2rtp, gao2021deep, cai2023m2g4rtp}. In this way, if the route prediction is wrong, the accuracy of the time prediction can also be affected. Moreover, the error in route prediction could accumulate and trim down the performance of time prediction. 
\par \textbf{Lack of public data and Benchmark.} Lastly, there is still a lack of public data in this area. Most existing conduct experiments with private data. For example, OSquare utilizes the data collected by Eleme, DeepRoute utilizes the data collected by Cainiao, and ILRoute uses the data by Meituan. Such data settings lack transparency and make it hard to reproduce the results. Although one recent work \cite{wu2023lade} proposes a publicly-available dataset (named LaDe) from the last-mile delivery, there is still a lack of publicly available and widely accepted datasets and benchmarks, which puts a hurdle to the development of this area. 

\subsection{Future Directions}
\par  There exist some interesting directions for future research. 

\par \textbf{More efficient decoding technology for route prediction.} One possible future direction is to develop a more efficient decoding mechanism. There can be two ways to achieve this goal. The first is from the perspective of model architecture, a non-autoregressive decoder can be explored that can generate multiple outputs at once. And the second is from the perspective of model compression \cite{choudhary2020comprehensive}, a more lightweight model can be explored to accelerate the inference speed when applied in the real system.

\par \textbf{Modeling of the road network.} As elaborated in the limitation, the road network contains abundant spatial information, and it is also the geographical space where the workers finish their tasks. Therefore, one future direction is to model the road network in the model design. For example, cast the problem instance into the road network, and reformulate the RTP problem based on the road network. In this background, how to effectively incorporate the information in the road network to boost the RTP performance, would be a quite challenging and promising direction.

\par \textbf{Modeling of the joint distribution of route and time.} Current route and time prediction models usually treat route prediction and time prediction separately. They either output the route and time in a two-step way \cite{wen2023enough}, or calculate the route and time by two modules \cite{gao2021deep}. However, route and time are actually strongly correlated with each other.  In light of this, it remains a potential future direction to develop more effective models that can represent them in a single unified manner where the two items are considered as a whole, and can capture the joint distribution of the route and time by leveraging abundant spatial-temporal context.

\par \textbf{Consideration of different route constraints.} Current models have already explored the case with the pickup-then-delivery route constraints. However, many different route constraints exist in different scenarios, such as capacity constraints \cite{gutierrez2020iot, zheng2019multi}, and first-in-last-out constraints \cite{zheng2007two, yan2020hybrid}. Specific model design may be required to handle different route constraints. It would be another future direction to develop models for route and time prediction problems under different route constraints, so that the solution can better align with the real scenarios. 

\par \textbf{Probabistic RTP forecasting.} Existing efforts all study the scenario where the point estimation is conducted. Specifically, for time prediction, only one point estimation is given by the model. And for route prediction, they usually give one route estimation. However, such kind of predictions cannot qualify the uncertainty of the prediction, which is crucial for downstream tasks that require risk assessment or decision-making under uncertainty \cite{salinas2020deepar, rasul2021autoregressive}. In future work, an interesting as well as challenging direction is to develop models that can give probabilistic forecasting. For example, proposing models that can evaluate the uncertainty of the time prediction. As for route prediction, one can develop models that can predict multiple possible routes at once. Furthermore, it would be more challenging and promising to conduct the joint probabilistic route and time forecasting.

\section{Conclusion} \label{sec:conclusion}
\par The realm of instant delivery services is experiencing unprecedented growth, largely attributed to its profound impact on enhancing daily living. This paper delves into the route and time prediction (RTP) problem in instant delivery—an essential component for implementing an intelligent delivery platform that has a significant influence on customer satisfaction and operational cost. In this paper, we present the first systematic survey of deep neural networks tailored for service route and time prediction in instant delivery. Specifically, we first introduce the problem, commonly used metrics, and propose a novel taxonomy to classify the existing models from three perspectives. Then, we elaborate on the details of the models in each class, focusing on their motivation and model architecture. At last, we introduce the limitations and discuss the potential future direction in this field. We believe that this review fills the gap in RTP research and ignites further research interest in the RTP problem.

\newpage

\bibliographystyle{IEEEtran}
\bibliography{IEEEabrv, ./img/my_bib.bib}

\end{document}